\documentclass[11pt]{article}

\usepackage[final]{acl}

\usepackage{times}
\usepackage{latexsym}

\usepackage[T1]{fontenc}

\usepackage[utf8]{inputenc}

\usepackage{microtype}

\usepackage{inconsolata}

\usepackage{graphicx}

\usepackage{amsmath,amsfonts,bm}

\def\eqref#1{equation~\ref{#1}}

\def\1{\bm{1}}

\DeclareMathAlphabet{\mathsfit}{\encodingdefault}{\sfdefault}{m}{sl}
\SetMathAlphabet{\mathsfit}{bold}{\encodingdefault}{\sfdefault}{bx}{n}

\usepackage{hyperref}
\usepackage{url}

\usepackage{times}
\usepackage{latexsym}
\usepackage{booktabs}
\usepackage{multirow}
\usepackage{tabularx}
\usepackage{amsmath}
\usepackage{wrapfig}
\usepackage{stfloats}
\usepackage{xcolor}
\usepackage{color, colortbl}
\definecolor{Gray}{gray}{0.9}
\definecolor{blizzardblue}{rgb}{0.67, 0.9, 0.93}
\definecolor{aliceblue}{rgb}{0.94, 0.97, 1.0}
\definecolor{babyblueeyes}{rgb}{0.63, 0.79, 0.95}
\definecolor{beaublue}{rgb}{0.74, 0.83, 0.9}
\definecolor{azure(web)(azuremist)}{rgb}{0.94, 1.0, 1.0}
\usepackage{mdframed}
\usepackage{algorithm}
\usepackage{algpseudocode}
\usepackage[english]{babel}
\usepackage{amsthm}

\usepackage{nicematrix}

\usepackage{hyperref}
\usepackage{url}
\usepackage{natbib}
\usepackage{times}
\usepackage{microtype}
\usepackage{adjustbox}
\usepackage{graphicx}
\usepackage{amsmath}
\usepackage{booktabs}
\usepackage{multirow}
\usepackage{float}
\usepackage{algorithm}
\usepackage{algpseudocode}
\usepackage{adjustbox}
\usepackage{subcaption}
\usepackage{enumitem}
\usepackage{fontawesome5}

\usepackage[most]{tcolorbox}
\tcbset{aibox/.style={
    breakable,
    break at=\maxdimen,
    fontlower=\scriptsize, 
}}

\newtcolorbox{AIbox}[2][]{aibox,title={#2},#1}

\title{AutoGraph-R1: End-to-End Reinforcement Learning for Knowledge Graph Construction}

\author{Hong Ting Tsang$^1$, Jiaxin Bai$^{1,}$\thanks{Corresponding author.}, Haoyu Huang$^1$, Qiao Xiao$^2$, 
\\ \textbf{Tianshi Zheng$^1$, Baixuan Xu$^1$, Shujie Liu$^3$, Yangqiu Song$^{1}$}\\
  $^1$The Hong Kong University of Science and Technology\\
    $^2$Cornell University\\
    $^3$Microsoft Research\\
  \texttt{\{httsangaj, jbai\}@connect.ust.hk}, \texttt{yqsong@cse.ust.hk} \\
\texttt{shujliu@microsoft.com} 
}

\begin{document}
\maketitle
\begin{abstract}
Building effective knowledge graphs (KGs) for Retrieval-Augmented Generation (RAG) is pivotal for advancing question answering (QA) systems. However, its effectiveness is hindered by a fundamental disconnect: the knowledge graph (KG) construction process is decoupled from its downstream application, yielding suboptimal graph structures. To bridge this gap, we introduce AutoGraph-R1, the first framework to directly optimize KG construction for task performance using Reinforcement Learning (RL). AutoGraph-R1 trains an LLM constructor by framing graph generation as a policy learning problem, where the reward is derived from the graph's functional utility in a RAG pipeline. We design two novel, task-aware reward functions, one for graphs as knowledge carriers and another as knowledge indices. Across multiple QA benchmarks, \textbf{AutoGraph-R1} consistently enables graph RAG methods to achieve significant performance gains over using task-agnostic baseline graphs. Our work shows it is possible to close the loop between construction and application, shifting the paradigm from building intrinsically ``good'' graphs to building demonstrably ``useful'' ones.
\end{abstract}

\section{Introduction}

Retrieval-Augmented Generation (RAG) has become a cornerstone for enhancing Large Language Models (LLMs), enabling them to ground responses in external knowledge, reduce hallucinations, and incorporate real-time, domain-specific information \citep{gao2024retrievalaugmentedgenerationlargelanguage, peng2024graphretrievalaugmentedgenerationsurvey}. A particularly promising frontier is graph-based RAG, where LLMs leverage the structured nature of Knowledge Graphs (KGs) for complex data sensemaking and reasoning \citep{edge_local_2025}. The typical pipeline begins by constructing a KG from unstructured text, often using LLM-driven extractors and heuristics \citep{han_pive_2024, lairgi_itext2kg_2024, bai_autoschemakg_2025}, which then supports downstream question answering \citep{gutierrez_hipporag_2025, sun_think--graph_2024}.

\begin{figure*}
\vspace{-0.5in}
    \centering
    \includegraphics[width=0.75\linewidth]{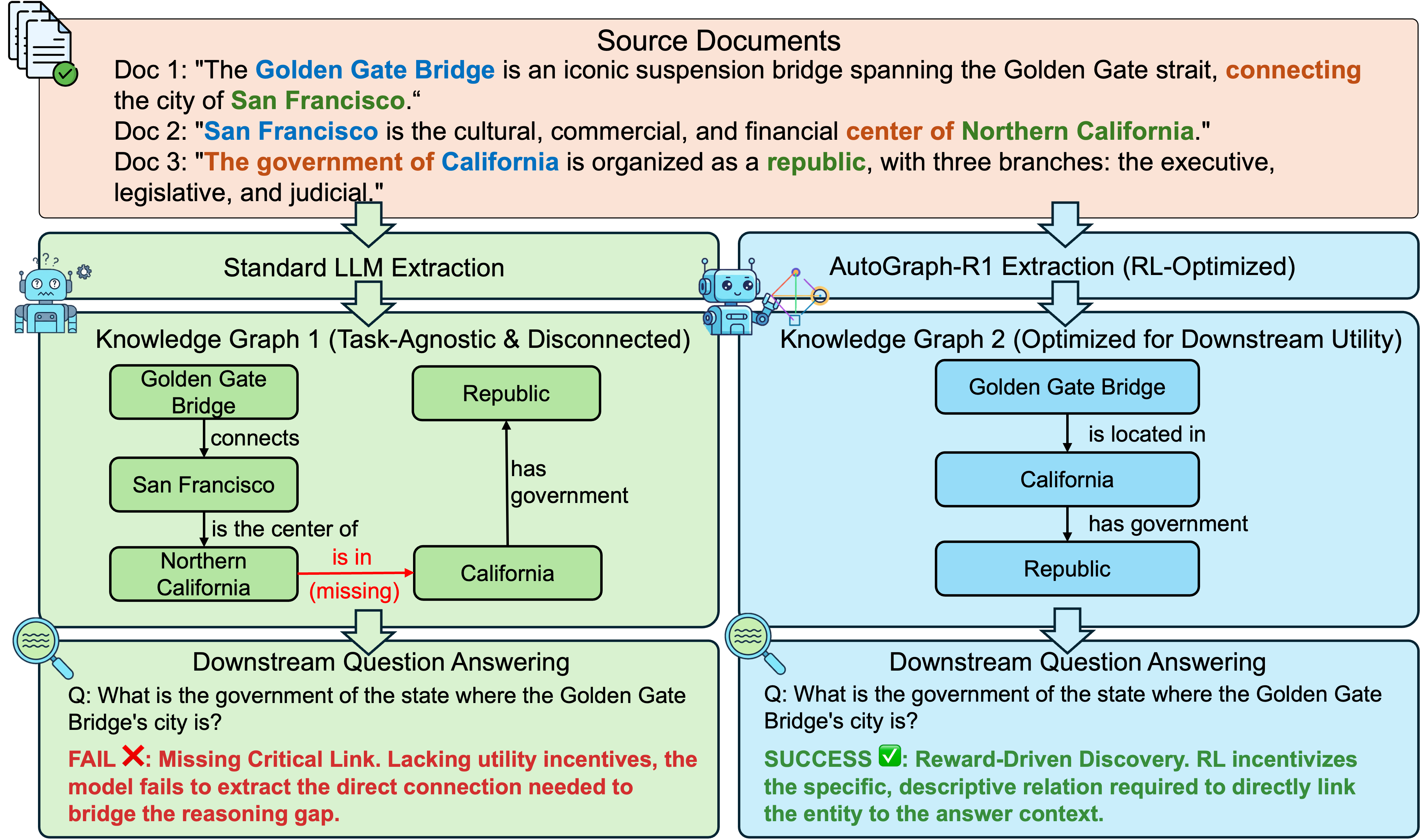}
    \caption{\textbf{Bridging the Gap Between KG Construction and Utility.} (A) Traditional extraction is task-agnostic and optimized for intrinsic metrics, often failing to capture implicit connections or diluting relevance signals with noise. (B) \textbf{AutoGraph-R1} aligns the graph structure with the specific downstream retriever. Driven by the reward, it learns to construct concrete reasoning pathways, or optimize graph connectivity to ensure robust signal propagation, delivering the specific structure required for the target task.}
    \label{fig:teaser}
\vspace{-0.1in}
\end{figure*}

Despite its potential, the prevailing graph-based RAG paradigm suffers from a fundamental disconnect: the process is split into two isolated phases. First, a construction phase, where a graph is built and evaluated on intrinsic metrics like precision and coverage \citep{huang_can_2025}. Second, an application phase, where this static graph is used for a downstream task. The critical flaw in this approach is that a ``good'' graph by intrinsic standards is not necessarily a ``useful'' one for the end task \citep{xue_knowledge_2022}. For instance, striving for exhaustive completeness often creates noisy, fragmented graphs where relevance signals are diluted or critical reasoning paths are broken, as illustrated in Figure \ref{fig:teaser}.

This disconnect persists because closing the loop between downstream performance and graph construction is technically challenging. The construction process generating discrete (subject, predicate, object) triples is inherently non-differentiable. Consequently, standard gradient-based optimization cannot backpropagate performance signals from a downstream task, such as question answering accuracy, to guide the graph generation model. The graph, once built, cannot learn from its failures. To bridge this gap, we employ Reinforcement Learning (RL). While prior work has used RL to refine retrieval over search tools or improve query reformulation \citep{jin_search-r1_2025, jiang_s3_2025, luo_graph-r1_2025}, our work is the first to leverage RL to directly optimize the KG construction process itself. We introduce AutoGraph-R1, a framework that fine-tunes an LLM-based graph generator by optimizing for downstream task performance. As shown in Figure \ref{fig:workflow}, the graph generation model learns a construction policy from raw text. The utility of the generated graph is then evaluated in a downstream RAG pipeline, yielding a reward signal. This task-aware reward is used to update the generator's parameters via policy gradient methods, guiding it to produce graphs that are not just factually accurate but functionally optimal, for instance, by creating valid paths that facilitate complex reasoning.

We specifically focus on question answering over benchmark datasets requiring multi-document reasoning \citep{yang_hotpotqa_2018, trivedi_musique_2022, mallen_when_2023}. Our framework's reward function is based on the utility of the resulting graph, evaluating whether it serves as an effective knowledge index for retrieving useful text chunks or provides subgraphs that directly support the reasoning process. By designing these task-aware rewards, we directly align the objectives of KG construction with end-task performance.

In summary, our contributions are: 
\begin{itemize} 
    \item We introduce AutoGraph-R1, a novel RL framework that directly optimizes knowledge graph construction for downstream utility, bridging the critical gap between graph quality and task performance. 
    \item We design and implement task-aware reward functions that successfully align KG structure with the demands of complex reasoning tasks, compelling the model to build functionally superior graphs. 
    \item Through extensive experiments, we demonstrate that integrating AutoGraph-R1's graphs into a state-of-the-art RAG pipeline yields significant performance gains on QA benchmarks, validating that RL-driven graph construction improves downstream task utility. 
\end{itemize}

\section{Related Work}
\subsection{Graph-based Retrieval-Augmented Generation}
Large Language Models (LLMs), despite demonstrating strong reasoning capabilities \citep{deepseek-ai_deepseek-r1_2025}, remain susceptible to factual hallucinations \citep{ji_survey_2023, huang_survey_2025} and knowledge incompleteness \citep{peng_study_2023}. Retrieval-Augmented Generation (RAG) \citep{lewis_retrieval-augmented_2021, gao_retrieval-augmented_2024} mitigates these issues by grounding LLMs in external knowledge sources, thereby improving factual accuracy and reasoning.
A burgeoning area of research extends RAG with graph-structured knowledge \citep{peng_graph_2024, xiang_when_2025, zhang_survey_2025, han_rag_2025}. In these pipelines, graphs serve two primary functions. First, as knowledge indices, where the graph organizes and connects raw text chunks, and its structural properties are leveraged for more sophisticated retrieval strategies \citep{liang_empowering_2024,liu_graphcoder_2024,zhang_multi-domain_2024,wang_knowledge_2023, li_graphreader_2024}. Methods like HippoRAG \citep{gutierrez_hipporag_2025, gutierrez_rag_2025} exemplify this by exploiting structural connections to access relevant information more effectively \citep{xiao_terag_2025}. Second, as knowledge carriers, where the graph itself is the primary information source, and the model reasons directly over recovered subgraphs \citep{shen_reason-align-respond_2025,liu_knowledge_2024}. This paradigm is adopted by approaches such as Think-on-Graph \citep{ma_think--graph_2025, sun_think--graph_2024}, SubgraphRAG \citep{li_simple_2025}, StructRAG \citep{li_structrag_2024}, and KnowGPT \citep{zhang_knowgpt_2024}.
\newline The construction of KGs has evolved from traditional rule-based systems like OpenIE \citep{angeli_leveraging_2015} to more flexible LLM-based pipelines such as PiVE \citep{han_pive_2024}, iText2KG \citep{lairgi_itext2kg_2024}, KGGEN \citep{mo_kggen_2025}, GraphRAG \citep{edge_local_2025}, LightRAG \citep{guo_lightrag_2025} and AutoSchemaKG \citep{bai_autoschemakg_2025}. While powerful, these LLM-driven methods typically generate a static graph based on fixed prompts or heuristics, often evaluated using intrinsic metrics. However, the optimal structure of a KG is highly dependent on variety of downstream applications \citep{gubanov_cancerkgorg_2024,wu_medical_2024,zhao_agentigraph_2024,he_enhancing_nodate,he_g-retriever_2024, liu_codexgraph_2024, bai_top_2025, huang_atlaskv_2026}. For instance, a graph acting as a text index may prioritize fine-grained partitioning, while one used for reasoning chains requires long-range connectivity \citep{jin_graph_2024,huang_ritek_2024}. This creates the disconnect we identified earlier: a graph built to be ``good'' in isolation may be functionally poor for a specific task. Our work addresses this gap by optimizing graph construction directly for downstream performance, a problem that, to our knowledge, has not been systematically investigated, despite progress in KG refinement and completion techniques \citep{chen_differentiable_nodate, chen_entity_2024, chen_neuro-symbolic_2024, zhang_contrastive_2022, dong_active_2023}.

\subsection{Reinforcement Learning for Language Model Optimization}
Reinforcement learning (RL) \citep{kaelbling_reinforcement_1996} offers a powerful framework for optimizing the sequential decision-making capabilities of LLMs by enabling them to learn through environmental interaction and reward feedback \citep{kaufmann_survey_2024, xi_survey_2025}. As LLMs have become more powerful through fine-tuning, methodological advances—from Reinforcement Learning from Human Feedback (RLHF) \citep{ouyang_training_2022} to more scalable and cost-effective algorithms like Proximal Policy Optimization (PPO) \citep{schulman_proximal_2017}, Dynamic Sampling Policy Optimization (DAPO) \citep{yu_dapo_2025}, and Group Relative Policy Optimization (GRPO) \citep{deepseek-ai_deepseek-r1_2025}—have enabled successful applications in diverse domains, including open-domain retrieval and scientific discovery. \citep{zheng_automation_2025, yu_mathcalb-coder_2024, zhu_convsearch-r1_2025, shen_vlm-r1_2025}.

Prior to its widespread adoption for LLM alignment, RL had been explored for knowledge base tasks, such as bidirectional text-to-graph conversion \citep{dognin_regen_2021} and abductive reasoning over knowledge graphs \citep{bai_advancing_2024}. More recently, RL has proven effective in training LLMs to interact with external tools, such as search engines \citep{jin_search-r1_2025, jiang_s3_2025, li_towards_2025, zhang_web_2025, jiang_deepretrieval_2025}. Notably, frameworks like Graph-R1 \citep{luo_graph-r1_2025} have shown that RL can teach an LLM to effectively navigate graph-structured tools to improve retrieval. 
However, these works use RL to learn a policy for navigating or querying an existing knowledge source. Our approach is fundamentally different: we use RL to learn a policy for constructing the knowledge source from raw text. While work like \citep{zhang_collaborative_2025} utilizes RL to complete or refine existing KGs during reasoning, to our knowledge, this is the first application of RL to directly optimize end-to-end graph generation from corpora based on its measured utility in a downstream task. This distinction forms the motivation for AutoGraph-R1.

\section{Preliminaries}
\label{sec:preliminaries}

In this section, we formalize the key concepts underlying AutoGraph-R1, including knowledge graph construction, graph-based retrieval, and answer generation within a RAG pipeline.

\subsection{Knowledge Graph Construction}
We define a knowledge graph (KG) as a directed, labeled graph $\mathcal{G} = (\mathcal{V}, \mathcal{E}, \mathcal{R})$, constructed from a set of documents $\mathcal{D}$. Here, $\mathcal{V}$ is the set of nodes, $\mathcal{R}$ the set of relation types, and $\mathcal{E} \subseteq \mathcal{V} \times \mathcal{R} \times \mathcal{V}$ the set of edges represented as triples $(s, r, o)$. Nodes $s,o \in \mathcal{V}$ may correspond to entities, events, or concepts, and $r \in \mathcal{R}$ denotes a relation type. Following prior work \citep{zhang_survey_2025}, we consider two principal configurations for the graph's role.

\textit{\textbf{Graphs as Knowledge Carriers}} In this configuration, the graph serves as a self-contained knowledge base. It consists of factual triples $(s, r, o) \in \mathcal{E}$, where nodes $s, o \in \mathcal{V}$ are entities and $r \in \mathcal{R}$ is the relation. These triples act as discrete, structured knowledge units that are retrieved and processed directly by the downstream model.

\textit{\textbf{Graphs for Knowledge Indexing}} Alternatively, the graph functions as a structured index over the raw document corpus $\mathcal{D}$. Nodes are augmented with pointers to text spans, denoted $\tau(v)$. Formally, the node set can be partitioned into entity nodes $\mathcal{V}_e$ and document or chunk nodes $\mathcal{V}_d$, such that $\mathcal{V} = \mathcal{V}_e \cup \mathcal{V}_d$. This hybrid structure allows the graph to guide retrieval not only of structured facts but also of the original, unstructured text passages.
\subsection{Retrieval Module}
Given a query $q$, the goal of the retrieval module is to produce a set of evidence units $\mathcal{C}(q)$ that will be passed to the LLM for answer generation. We consider two complementary retrieval strategies.

\textit{\textbf{Graph Knowledge Retriever}} This retriever, denoted $\mathcal{R}_{\text{graph}}$, operates directly on the graph structure. Given a query $q$, it identifies relevant components such as individual triples, multi-hop paths, or entire subgraphs. Formally, we define its output as a set of structured evidence $\mathcal{P}(q) = \mathcal{R}_{\text{graph}}(q, \mathcal{G})$. The elements of $\mathcal{P}(q)$ are then linearized into text to serve as context for the LLM.

\textit{\textbf{Graph-based Text Retriever}} This retriever, denoted $\mathcal{R}_{\text{text}}$, uses the graph as an index to find relevant text passages from the source corpus. It leverages graph connectivity to identify promising document nodes. Formally, $\mathcal{R}_{\text{text}}(q, \mathcal{G}) \mapsto \mathcal{T}(q)$, where $\mathcal{T}(q) \subseteq \{\tau(v) \mid v \in \mathcal{V}_d\}$ is a set of raw text passages linked from document nodes.

\subsection{Answer Generation}
The final answer generation step uses a large language model $\pi^{ans}$ to synthesize an answer $\hat{y}$ from the query $q$ and the retrieved evidence $\mathcal{C}(q)$. The evidence context $\mathcal{C}(q)$ is composed of either the linearized graph structures $\mathcal{P}(q)$ from the graph knowledge retriever or the text passages $\mathcal{T}(q)$ from the graph-based text retriever. The final answer $\hat{y}$ is generated by conditioning the LLM on the query and evidence: $\hat{y} = \pi^{ans}(q, \mathcal{C}(q))$. This unified framework allows our optimization process to apply to both types of graph construction, directly linking the structure of $\mathcal{G}$ to its utility in the final QA-task.

\section{AutoGraph-R1}

\subsection{RL for Graph Construction}

\textbf{AutoGraph-R1}, an end-to-end reinforcement learning (RL) framework that directly optimizes knowledge graph (KG) construction with downstream task performance as the reward signal. The framework unifies two common graph-augmented retrieval paradigms: \emph{Graph RAG} (retrieval over entity triples) and \emph{Graph Text RAG} (retrieval over text nodes through graph index).  

\begin{figure*}
\vspace{-0.5
in}
    \centering
    \includegraphics[width=0.8\linewidth]{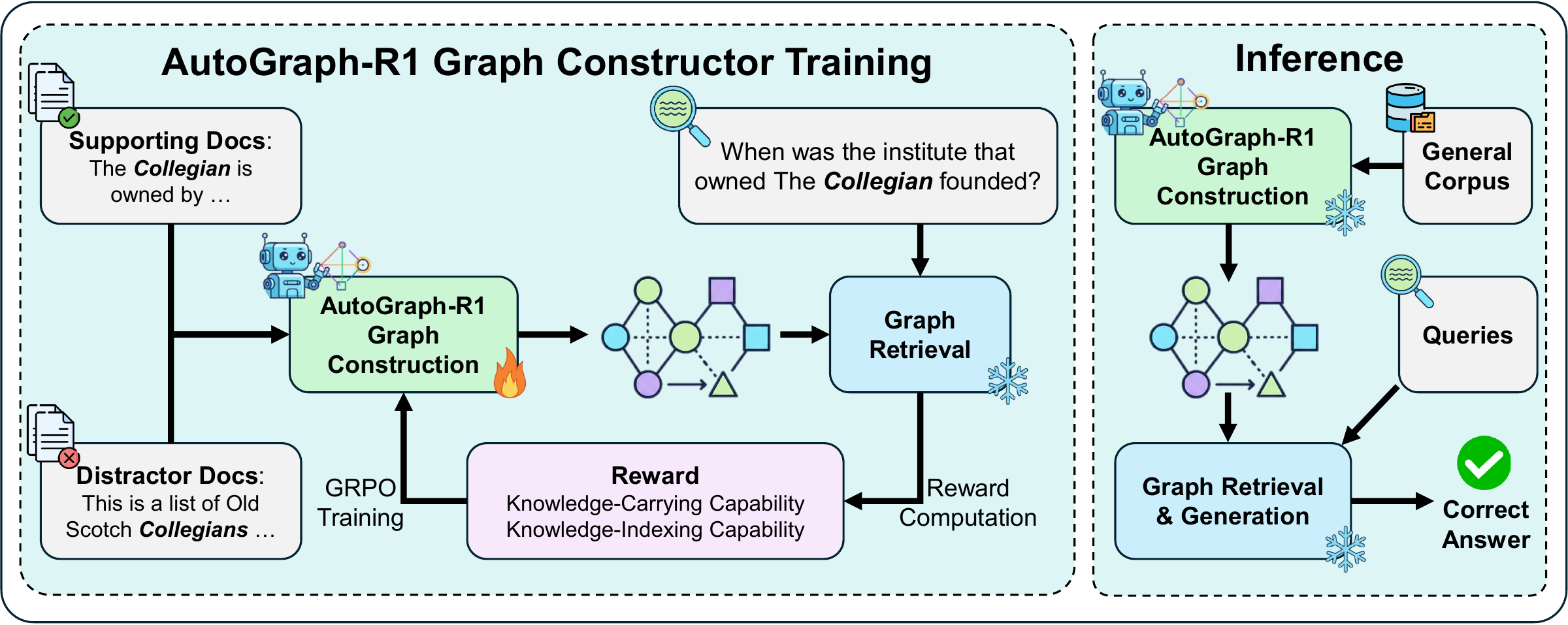}
    \caption{Overview of the AutoGraph-R1 Framework. AutoGraph-R1 optimizes knowledge graph construction for downstream utility using reinforcement learning. During the training phase (left), a graph constructor is fine-tuned with GRPO. The reward signal is derived from the performance of a graph retriever on the generated KG, directly measuring the graph's functional quality. During the inference phase (right), the trained constructor is used to build a large-scale KG from a general corpus, which then serves a downstream graph-based RAG system.}
    \label{fig:workflow}
\vspace{-0.15in}
\end{figure*}

As shown in Figure~\ref{fig:workflow}, AutoGraph-R1 consists of three components:  
(1) a KG construction policy model $\pi_{\theta}^{KG}$, instantiated as a large language model (LLM), which maps a list of documents $\mathcal{D}$ into a graph $\mathcal{G}$;  
(2) a frozen RAG server with a fixed answer generator $\pi^{ans}$, which retrieves from $\mathcal{G}$ and produces an answer $\hat{y}$ to the input query $q$;  
(3) a task-specific reward function $R(q, \hat{y}, y, \mathcal{G})$ that evaluates how well the constructed graph supports QA, where $y$ is the gold answer.

\textit{\textbf{Task-Aware Training Loop}} A central design choice in AutoGraph-R1, inspired by s3 \citep{jiang_s3_2025}, is to freeze the retrieval module while the KG construction policy $\pi_{\theta}^{KG}$ adapts. During training, each sample $(q, y, \mathcal{D}_q)$ comprising a query, a gold answer, and relevant documents, triggers a full end-to-end loop of KG construction, retrieval, and answer generation. Crucially, the definition of a "useful" graph is contingent on the retrieval paradigm. We therefore tailor the training process for two distinct scenarios, aligning the KG's structure with its intended function.

\textit{\textbf{Training with a Graph Knowledge Retriever.}} When the KG acts as a knowledge carrier, retrieval quality is measured by its ability to provide a self-contained, structured context for reasoning. To isolate the impact of graph structure, we employ a simple subgraph retriever. Given a query $q$, we extract its named entities to serve as anchors and retrieve the $n$-hop neighborhood surrounding them to form the context $\mathcal{P}(q)$. This design intentionally bypasses dense vector similarity, forcing the reward signal to reflect the graph's relational completeness and structural integrity. The policy is rewarded for creating graphs where the correct answer is directly \emph{deducible} from the retrieved subgraph.

\textit{\textbf{Training with a Graph-based Text Retriever.}} When the KG serves as a knowledge index, its utility is determined by how well it guides the retriever to relevant text passages. We adapt the HippoRAG-2 retriever \citep{gutierrez_rag_2025} for this purpose. First, candidate triples $(s,r,o)$ are selected based on embedding similarity to the query $q$ (using Qwen-3-0.6B). However, unlike the original method, we then use \emph{only} these triple-level similarities to initialize a Personalized PageRank algorithm over the constructed graph $\mathcal{G}$. This process propagates relevance scores through the graph structure, ultimately identifying text nodes that are structurally connected to the most pertinent facts, from which the top-$N$ passages are returned. The policy is therefore incentivized to build graphs where structural connectivity, not just semantic similarity, is a reliable signal for identifying crucial evidence passages, including both direct and complementary information that might otherwise be overlooked.

\medskip
In both scenarios, by freezing the retriever and aligning the reward with its specific mechanism, AutoGraph-R1 ensures the graph construction policy learns to produce graphs that are optimized for a given downstream retrieval strategy.

\subsection{Reward Design for Functional Graph Construction}
A primary challenge in optimizing this end-to-end pipeline is the sparse and indirect nature of the learning signal, where a single reward is given after a long sequence of construction actions. This creates a severe credit assignment problem and places a heavy burden on the quality of the reward signal itself. While the final answer's F1 score has been explored as a reward in prior work \citep{jin_search-r1_2025}, we find its properties make it a challenging choice for our goal of guiding graph construction. The F1 score is brittle; minor phrasing variations in the LLM output can cause swings in the metric, an issue that persists even with deterministic decoding. This instability results in a noisy reward that can impede or destabilize policy optimization. Our approach is therefore motivated by the need to design more direct and stable, task-specific rewards better suited to our problem.

To overcome these challenges, we design two distinct reward functions that provide a more direct and stable learning signal by measuring the functional utility of the graph for a specific retrieval task.

\textbf{\textit{Graph Knowledge Retriever}} Retrieval operates on subgraphs or relation paths. The key requirement is that the gold answer $y$ should be \emph{deducible} from $\mathcal{G}$. An answer $y$ is considered deducible from $\mathcal{G}$ if the retrieved triples or subgraphs contain sufficient relational information to logically infer the gold answer $y$ for a given query $q$, either directly through explicit facts or indirectly through reasoning over connected triples. We therefore define a binary reward, the \textbf{Knowledge-Carrying Reward}, $R_C$, which measures the \emph{deducibility} of gold answer in the constructed KG. An external LLM judge is prompted with $(q, y, \mathcal{G})$ and determines whether $y$ can be deduced from the retrieved triples:
\begin{equation}
    R_{\text{C}}(q, y, \mathcal{G}) = \mathbb{I}\big[\textit{deducible}(q, y \mid \mathcal{G})\big]
\label{deducible_reward}
\end{equation}
\textbf{\textit{Graph-based Text Retriever}} Retrieval operates over the graph structure of the KG to locate relevant text passages. To enhance the \emph{knowledge-indexing capability}, we use \textbf{Knowledge-Indexing Reward}, $R_I$, as the reward function. This aligns with the fundamental objective of these retrievers by measuring the effectiveness of the retrieved passages in capturing the relevant information. The reward is defined as follows:
\begin{equation}
R_{\text{I}}(q, \mathcal{D}_{\text{gold}}, \mathcal{G}) = 
\frac{\left|\text{Top-}k(\mathcal{G}, q) \cap \mathcal{D}_{\text{gold}}\right|}
{\left|\mathcal{D}_{\text{gold}}\right|} 
\label{recall_reward}
\end{equation}
where $\mathcal{D}_{\text{gold}}$ denotes the gold passages for $q$, and $\text{Top-}k(\mathcal{G}, q)$ are the retrieved passages.

\subsection{GRPO for Graph Construction}

To optimize the knowledge graph (KG) constructor policy $\pi_\theta^{KG}$, we employ \textbf{Group-Relative Policy Optimization (GRPO)} \citep{shao_deepseekmath_2024}, a memory-efficient alternative to Proximal Policy Optimization (PPO). GRPO is well-suited for our LLM-based framework, as it eliminates the need for a separate value model by using a relative reward baseline derived from a group of sampled graph outputs. This approach reduces computational overhead and memory usage, enabling scalable training for large-scale graph construction tasks.

The GRPO objective updates $\pi_\theta^{KG}$ to favor graphs that maximize downstream QA performance, incorporating a clipping mechanism to ensure stable updates. We simplify the training procedure by removing the KL divergence term to lower the computational overhead and save memory usage without damaging the training \citep{liu_understanding_2025,hu_open-reasoner-zero_2025}. The objective is defined as:

\begin{equation}
\begin{split}
\mathcal{J}_{\text{GRPO}}(\theta) & \\
= \mathbb{E}_{\substack{\mathbf{s} \sim \mathcal{D} \\ \{\mathbf{a}_i\} \sim \pi_{\theta_{\text{old}}}}} & \bigg[ \frac{1}{G} \sum_{i=1}^{G} \sum_{t=1}^{|\mathbf{a}_i|} \min \Big( r_{i,t}(\theta) \hat{A}_i, \\
& \text{clip}(r_{i,t}(\theta), 1-\epsilon, 1+\epsilon) \hat{A}_i \Big) \bigg]
\end{split}
\end{equation}

\noindent Here, the policy $\pi_{\theta}^{KG}$ takes the document list $\mathbf{s}$ as input. In this equation, $\mathbf{s}$ corresponds to the input document list drawn from the training distribution $\mathcal{D}$. 

\textbf{Construction of Training Inputs ($\mathcal{D}$):} The composition of $\mathbf{s}$ is tailored to the specific retrieval paradigm to shape the learning signal. To create a challenging training environment for the \textbf{text retrieval scenario}, we implement a hard negative mining strategy. For each query in the training set (HotpotQA and Musique), we use a \texttt{Qwen3-8B} \citep{zhang_qwen3_2025} embedding model to identify the most semantically similar non-gold passage from the corpus, which is then added as a distractor. In contrast, for the \textbf{graph knowledge retriever scenario}, no distractors are used, as the primary objective is to optimize the informational completeness of the graph constructed from source documents, improving the knowledge-carrying capability of graph rather than its ability to filter irrelevant content. $\hat{A}_i = \frac{R_i-\mu_R}{\sigma_R}$ represents the Group-Relative Advantage for the entire graph $\mathbf{a}_i$, which is derived by normalizing its reward relative to the group's mean $\mu_R$ and standard deviation $\sigma_R$. The downstream reward signal, $R_i$, for the $i$-th graph sample is determined by the specific training setup: $R=R_{\text{C}}$ (Eq. \ref{deducible_reward}) when employing a graph knowledge retriever, or $R=R_{\text{I}}$ (Eq. \ref{recall_reward}) when a graph-based text retriever is used. $\epsilon$ is a small clipping hyperparameter that ensures stable updates by preventing the new policy from straying too far from the old policy.

\section{Experiments}
\label{sec:experiments}
Our experiments are designed to answer three primary research questions: \textbf{RQ1.} Does optimizing KG construction with a downstream task reward lead to better end-to-end RAG performance compared to standard, task-agnostic KG construction? \textbf{RQ2.} Is this performance improvement consistent across different graph-based RAG paradigms (i.e., graph as a \textit{knowledge carrier} vs. a \textit{knowledge index}) and across different model scales? \textbf{RQ3.} Does optimizing for downstream utility also improve the \textit{intrinsic quality} (e.g., factual precision and recall) of the graph, and how do different reward functions bias the final graph structure?
\subsection{Datasets and Corpora}

\textbf{Training Datasets} For the reinforcement learning phase, we utilize two multi-hop QA datasets: HotpotQA \citep{yang_hotpotqa_2018} and Musique \citep{trivedi_musique_2022}. These datasets provide the complex reasoning challenges necessary to train our policy model.

\textbf{Evaluation Datasets} For final RAG evaluation, we use a diverse set of five QA datasets, each comprising 1,000 samples. These include two general QA benchmarks, Natural Questions (NQ) \citep{kwiatkowski_natural_2019} and PopQA \citep{mallen_when_2023}, and three multi-hop QA benchmarks, HotpotQA \citep{yang_hotpotqa_2018}, 2WikiMultihopQA \citep{ho_constructing_2020}, and Musique \citep{trivedi_musique_2022}. These datasets enable a comprehensive evaluation of AutoGraph-R1 on downstream task-utility.

\textbf{Corpora.} For the general QA datasets (NQ, PopQA), the knowledge corpus is built from the introductory sections of the December 2021 Wikipedia dump \citet{izacard_atlas_2022}. For the multi-hop QA datasets, the corpus for each is constructed from the documents associated with its 1,000 evaluation samples, following the methodology in \citet{gutierrez_hipporag_2025}.

\subsection{Experiments Configs}
\label{sec:exp_config}
\textbf{Models} We experiment with fine-tuning both \texttt{Qwen2.5-3B} and \texttt{Qwen2.5-7B} \citep{qwen_qwen25_2025} as the KG construction model ($\pi_{\theta}^{KG}$). For all RAG evaluations, we use a frozen \texttt{Qwen2.5-7B} as the answer generation LLM. The \texttt{Qwen3-0.6B} model is used consistently for all embedding tasks in both training and evaluation.
\textbf{RL Training Configuration} We fine-tune the KG construction policy using the GRPO algorithm \citep{shao_deepseekmath_2024} on two \texttt{H100} GPUs. For each training sample, the policy iteratively generates triples for each document. The training setup is tailored to the retrieval paradigm. For the graph-based text retriever, the model operates on a fixed pool of 15 documents per query, retrieving the top-$N$ passages, where $N$ equals the number of gold supporting passages. For the graph knowledge retriever, we retrieve a 3-hop subgraph, to ensure that a sufficient context is always retrieved for reasoning.
\textbf{Evaluation Protocol} For evaluation, a KG is first constructed over the entire document corpus for each dataset. Then, depending on the type of retriever, the corresponding RAG is performed using this static graph. We report the final answer F1 score as the primary metric, consistent with prior work.

\subsection{Baselines}
To rigorously isolate the impact of downstream-aware optimization, we benchmark AutoGraph-R1 against a controlled, task-agnostic baseline using SOTA retrieval methods, while holding all other pipeline components constant.

\textbf{KG Construction Baseline} To benchmark the performance of our RL-optimized constructor, we establish a baseline using a zero-shot approach. Specifically, we construct the baseline knowledge graphs using \texttt{Qwen} models guided by the identical prompt used during fine-tuning. This standard, task-agnostic method allows us to isolate the impact to directly measure the gains attributable to our downstream-aware optimization, isolating the impact of the construction policy by excluding auxiliary refinement steps used in complex pipelines.

\textbf{RAG Method Baselines} We evaluate the KGs constructed by both AutoGraph-R1 and the zero-shot baseline using a suite of state-of-the-art RAG methods to measure their functional utility. For \textbf{graph knowledge retrieval}, where the graph itself is the source of information, we test three distinct approaches. First, we use \textbf{ToG} \citep{sun_think--graph_2024}, setting both the width and depth to 3. It performs beam search on the graph and uses the \texttt{Qwen3-0.6B} model for relation pruning to discover meaningful paths. Second, we employ a \textbf{Subgraph Retriever}, which first performs Named Entity Recognition (NER) on the query and then expands 1-hop from the identified entities to form the retrieval context. Third, we include a \textbf{Dense Triple Retriever}, which uses the \texttt{Qwen3-0.6B} model to retrieve triples based on the semantic similarity between their embeddings and the query embedding. For these methods, the top-10 retrieved paths or triples are used as context. For \textbf{graph-based text retrieval}, where the graph serves as an index over a text corpus, we use \textbf{HippoRAG} \citep{gutierrez_hipporag_2025} and \textbf{HippoRAG-2} \citep{gutierrez_rag_2025}. Both methods perform query-to-edge retrieval to identify relevant triples, which then seed a Personalized PageRank (PPR) algorithm over the KG to score and rank text passages. The top-5 ranked passages are then returned as evidence.

\subsection{Results and Analysis}

\begin{table*}[!ht]
\vspace{-0.2in}
\small
\centering
\caption{Comprehensive performance evaluation of AutoGraph-R1 across Qwen models. The table is divided into two sections: (Top) Graph Knowledge Retrievers evaluated on F1 score, and (Bottom) Graph-based Text Retrievers evaluated on both F1 score and Passage Recall@5 (R@5). AutoGraph-R1 (``Ours'') consistently outperforms the zero-shot baselines across all metrics and datasets.}
\label{tab:unified_results}
\resizebox{\textwidth}{!}{
\begin{NiceTabular}{@{}l|cc|cc|cc|cc|cc|cc@{}}
\toprule
\multirow{2}{*}{\textbf{Methods}}
  & \multicolumn{2}{c|}{\textbf{NQ*}}
  & \multicolumn{2}{c|}{\textbf{PopQA*}}
  & \multicolumn{2}{c|}{\textbf{HotpotQA}}
  & \multicolumn{2}{c|}{\textbf{2WikiMultihopQA}}
  & \multicolumn{2}{c|}{\textbf{Musique}}
  & \multicolumn{2}{c}{\textbf{Avg.}} \\
& \multicolumn{2}{c|}{F1}
& \multicolumn{2}{c|}{F1}
& \multicolumn{2}{c|}{F1}
& \multicolumn{2}{c|}{F1}
& \multicolumn{2}{c|}{F1}
& \multicolumn{2}{c}{F1} \\
\midrule

\multicolumn{13}{c}{\cellcolor[gray]{0.9}\textbf{Graph Knowledge Retrievers (F1 Only)}} \\
\midrule

\rowcolor[gray]{0.9}{\textit{Qwen2.5-3B}} \\
\midrule
Subgraph (Base)
  & \multicolumn{2}{c|}{26.43}
  & \multicolumn{2}{c|}{54.48}
  & \multicolumn{2}{c|}{39.02}
  & \multicolumn{2}{c|}{34.15}
  & \multicolumn{2}{c|}{13.82}
  & \multicolumn{2}{c}{33.58} \\
\rowcolor{beaublue}Subgraph (Ours)
  & \multicolumn{2}{c|}{\textbf{28.03}\textcolor{red}{$\uparrow_{1.60}$}}
  & \multicolumn{2}{c|}{\textbf{59.46}\textcolor{red}{$\uparrow_{4.98}$}}
  & \multicolumn{2}{c|}{\textbf{40.77}\textcolor{red}{$\uparrow_{1.75}$}}
  & \multicolumn{2}{c|}{\textbf{34.71}\textcolor{red}{$\uparrow_{0.56}$}}
  & \multicolumn{2}{c|}{\textbf{15.13}\textcolor{red}{$\uparrow_{1.31}$}}
  & \multicolumn{2}{c}{\textbf{35.62}\textcolor{red}{$\uparrow_{2.04}$}} \\
Triples (Base)
  & \multicolumn{2}{c|}{30.53}
  & \multicolumn{2}{c|}{51.67}
  & \multicolumn{2}{c|}{40.76}
  & \multicolumn{2}{c|}{32.18}
  & \multicolumn{2}{c|}{17.81}
  & \multicolumn{2}{c}{34.58} \\
\rowcolor{beaublue}Triples (Ours)
  & \multicolumn{2}{c|}{\textbf{33.67}\textcolor{red}{$\uparrow_{3.14}$}}
  & \multicolumn{2}{c|}{\textbf{56.76}\textcolor{red}{$\uparrow_{5.09}$}}
  & \multicolumn{2}{c|}{\textbf{46.94}\textcolor{red}{$\uparrow_{6.18}$}}
  & \multicolumn{2}{c|}{\textbf{36.09}\textcolor{red}{$\uparrow_{3.91}$}}
  & \multicolumn{2}{c|}{\textbf{21.41}\textcolor{red}{$\uparrow_{3.60}$}}
  & \multicolumn{2}{c}{\textbf{38.97}\textcolor{red}{$\uparrow_{4.39}$}} \\
ToG (Base)
  & \multicolumn{2}{c|}{26.32}
  & \multicolumn{2}{c|}{54.92}
  & \multicolumn{2}{c|}{41.77}
  & \multicolumn{2}{c|}{43.54}
  & \multicolumn{2}{c|}{18.21}
  & \multicolumn{2}{c}{36.95} \\
\rowcolor{beaublue}ToG (Ours)
  & \multicolumn{2}{c|}{\textbf{29.27}\textcolor{red}{$\uparrow_{2.95}$}}
  & \multicolumn{2}{c|}{\textbf{61.40}\textcolor{red}{$\uparrow_{6.48}$}}
  & \multicolumn{2}{c|}{\textbf{44.56}\textcolor{red}{$\uparrow_{2.79}$}}
  & \multicolumn{2}{c|}{\textbf{49.33}\textcolor{red}{$\uparrow_{5.79}$}}
  & \multicolumn{2}{c|}{\textbf{18.42}\textcolor{red}{$\uparrow_{0.21}$}}
  & \multicolumn{2}{c}{\textbf{40.60}\textcolor{red}{$\uparrow_{3.65}$}} \\
\midrule

\rowcolor[gray]{0.9}{\textit{Qwen2.5-7B}} \\
\midrule
Subgraph (Base)
  & \multicolumn{2}{c|}{28.07}
  & \multicolumn{2}{c|}{55.43}
  & \multicolumn{2}{c|}{41.66}
  & \multicolumn{2}{c|}{33.97}
  & \multicolumn{2}{c|}{15.24}
  & \multicolumn{2}{c}{34.87} \\
\rowcolor{beaublue}Subgraph (Ours)
  & \multicolumn{2}{c|}{\textbf{28.54}\textcolor{red}{$\uparrow_{0.47}$}}
  & \multicolumn{2}{c|}{\textbf{60.94}\textcolor{red}{$\uparrow_{5.51}$}}
  & \multicolumn{2}{c|}{\textbf{43.59}\textcolor{red}{$\uparrow_{1.93}$}}
  & \multicolumn{2}{c|}{\textbf{37.43}\textcolor{red}{$\uparrow_{3.46}$}}
  & \multicolumn{2}{c|}{\textbf{15.65}\textcolor{red}{$\uparrow_{0.41}$}}
  & \multicolumn{2}{c}{\textbf{37.23}\textcolor{red}{$\uparrow_{2.36}$}} \\
Triples (Base)
  & \multicolumn{2}{c|}{33.26}
  & \multicolumn{2}{c|}{55.56}
  & \multicolumn{2}{c|}{44.99}
  & \multicolumn{2}{c|}{35.57}
  & \multicolumn{2}{c|}{20.43}
  & \multicolumn{2}{c}{37.96} \\
\rowcolor{beaublue}Triples (Ours)
  & \multicolumn{2}{c|}{\textbf{33.98}\textcolor{red}{$\uparrow_{0.72}$}}
  & \multicolumn{2}{c|}{\textbf{58.02}\textcolor{red}{$\uparrow_{2.46}$}}
  & \multicolumn{2}{c|}{\textbf{48.28}\textcolor{red}{$\uparrow_{3.29}$}}
  & \multicolumn{2}{c|}{\textbf{36.04}\textcolor{red}{$\uparrow_{0.47}$}}
  & \multicolumn{2}{c|}{\textbf{20.56}\textcolor{red}{$\uparrow_{0.13}$}}
  & \multicolumn{2}{c}{\textbf{39.38}\textcolor{red}{$\uparrow_{1.42}$}} \\
ToG (Base)
  & \multicolumn{2}{c|}{25.59}
  & \multicolumn{2}{c|}{57.53}
  & \multicolumn{2}{c|}{43.93}
  & \multicolumn{2}{c|}{46.03}
  & \multicolumn{2}{c|}{18.46}
  & \multicolumn{2}{c}{38.31} \\
\rowcolor{beaublue}ToG (Ours)
  & \multicolumn{2}{c|}{\textbf{29.36}\textcolor{red}{$\uparrow_{3.77}$}}
  & \multicolumn{2}{c|}{\textbf{62.85}\textcolor{red}{$\uparrow_{5.32}$}}
  & \multicolumn{2}{c|}{\textbf{44.68}\textcolor{red}{$\uparrow_{0.75}$}}
  & \multicolumn{2}{c|}{\textbf{50.20}\textcolor{red}{$\uparrow_{4.17}$}}
  & \multicolumn{2}{c|}{\textbf{19.31}\textcolor{red}{$\uparrow_{0.85}$}}
  & \multicolumn{2}{c}{\textbf{41.28}\textcolor{red}{$\uparrow_{2.97}$}} \\
\midrule\midrule

\multicolumn{13}{c}{\cellcolor[gray]{0.9}\textbf{Graph-based Text Retrievers (F1 \& Recall@5)}} \\
\midrule
& F1 & R@5 & F1 & R@5 & F1 & R@5 & F1 & R@5 & F1 & R@5 & F1 & R@5 \\
\midrule

\rowcolor[gray]{0.9}{\textit{Qwen2.5-3B}} \\
\midrule
HippoRAG (Base)   & 36.28 & 79.50 & 65.55 & 92.10 & 53.22 & 68.41 & 48.97 & 70.84 & 27.44 & 46.50 & 46.29 & 71.47 \\
\rowcolor{beaublue}HippoRAG (Ours)   & \textbf{38.28}\textcolor{red}{$\uparrow_{2.00}$} & \textbf{93.00}\textcolor{red}{$\uparrow_{13.5}$} & \textbf{65.93}\textcolor{red}{$\uparrow_{0.38}$} & \textbf{95.60}\textcolor{red}{$\uparrow_{3.50}$} & \textbf{55.39}\textcolor{red}{$\uparrow_{2.17}$} & \textbf{68.82}\textcolor{red}{$\uparrow_{0.41}$} & \textbf{51.69}\textcolor{red}{$\uparrow_{2.72}$} & \textbf{74.02}\textcolor{red}{$\uparrow_{3.18}$} & \textbf{28.11}\textcolor{red}{$\uparrow_{0.67}$} & \textbf{47.65}\textcolor{red}{$\uparrow_{1.15}$} & \textbf{47.88}\textcolor{red}{$\uparrow_{1.59}$} & \textbf{75.82}\textcolor{red}{$\uparrow_{4.35}$} \\
HippoRAG2 (Base)  & 35.88 & 82.20 & 65.02 & 92.20 & 53.70 & 70.06 & 50.98 & 73.49 & 25.70 & 46.93 & 46.25 & 72.98 \\
\rowcolor{beaublue}HippoRAG2 (Ours)  & \textbf{38.45}\textcolor{red}{$\uparrow_{2.57}$} & \textbf{94.00}\textcolor{red}{$\uparrow_{11.8}$} & \textbf{66.23}\textcolor{red}{$\uparrow_{1.21}$} & \textbf{95.40}\textcolor{red}{$\uparrow_{3.20}$} & \textbf{56.28}\textcolor{red}{$\uparrow_{2.58}$} & \textbf{71.21}\textcolor{red}{$\uparrow_{1.15}$} & \textbf{52.80}\textcolor{red}{$\uparrow_{1.82}$} & \textbf{76.42}\textcolor{red}{$\uparrow_{2.93}$} & \textbf{27.93}\textcolor{red}{$\uparrow_{2.23}$} & \textbf{49.13}\textcolor{red}{$\uparrow_{2.20}$} & \textbf{48.34}\textcolor{red}{$\uparrow_{2.09}$} & \textbf{77.23}\textcolor{red}{$\uparrow_{4.25}$} \\
\midrule
\rowcolor[gray]{0.9}{\textit{Qwen2.5-7B}} \\
\midrule
HippoRAG (Base)   & 37.16 & 93.30 & 65.95 & 92.90 & 55.50 & 69.97 & 53.01 & 72.87 & 26.03 & 47.19 & 47.53 & 75.24 \\
\rowcolor{beaublue}HippoRAG (Ours)   & \textbf{38.80}\textcolor{red}{$\uparrow_{1.64}$} & \textbf{94.30}\textcolor{red}{$\uparrow_{1.00}$} & \textbf{67.85}\textcolor{red}{$\uparrow_{1.90}$} & \textbf{95.80}\textcolor{red}{$\uparrow_{2.90}$} & \textbf{57.19}\textcolor{red}{$\uparrow_{1.69}$} & \textbf{71.40}\textcolor{red}{$\uparrow_{1.43}$} & \textbf{53.60}\textcolor{red}{$\uparrow_{0.59}$} & \textbf{76.16}\textcolor{red}{$\uparrow_{3.29}$} & \textbf{26.97}\textcolor{red}{$\uparrow_{0.94}$} & \textbf{48.44}\textcolor{red}{$\uparrow_{1.25}$} & \textbf{48.88}\textcolor{red}{$\uparrow_{1.35}$} & \textbf{77.22}\textcolor{red}{$\uparrow_{1.98}$} \\
HippoRAG2 (Base)  & 37.02 & 94.10 & 65.74 & 92.80 & 57.08 & 72.03 & 54.99 & 75.98 & 26.77 & 48.55 & 48.32 & 76.69 \\
\rowcolor{beaublue}HippoRAG2 (Ours)  & \textbf{38.68}\textcolor{red}{$\uparrow_{1.66}$} & \textbf{95.00}\textcolor{red}{$\uparrow_{0.90}$} & \textbf{67.72}\textcolor{red}{$\uparrow_{1.97}$} & \textbf{96.30}\textcolor{red}{$\uparrow_{3.50}$} & \textbf{58.98}\textcolor{red}{$\uparrow_{1.90}$} & \textbf{73.61}\textcolor{red}{$\uparrow_{1.58}$} & \textbf{56.46}\textcolor{red}{$\uparrow_{1.47}$} & \textbf{78.66}\textcolor{red}{$\uparrow_{2.68}$} & \textbf{27.18}\textcolor{red}{$\uparrow_{0.41}$} & \textbf{49.23}\textcolor{red}{$\uparrow_{0.68}$} & \textbf{49.80}\textcolor{red}{$\uparrow_{1.48}$} & \textbf{78.56}\textcolor{red}{$\uparrow_{1.87}$} \\
\bottomrule
\end{NiceTabular}}
\end{table*}

\textbf{\textit{AutoGraph-R1 consistently improves RAG performance across different paradigms, models, and scales.}} Our primary finding is that optimizing KG construction for downstream utility leads to significant end-to-end F1 score improvements over a standard zero-shot constructor, and this holds true for both Qwen and Llama (see Appendix \ref{sec:appendix_llama_result}) model families. As shown in Table \ref{tab:unified_results}, when the KG acts as a \textit{knowledge carrier}, our method yields substantial average F1 gains, up to +4.39 (Qwen-3B) and +2.97 (Qwen-7B). Similarly, when the KG is a \textit{knowledge index}, performance increases by up to +2.09 (Qwen-3B) and +1.48 (Qwen-7B) average F1 points. This robust improvement across different models and RAG paradigms confirms that task-aware optimization is broadly effective.

\textbf{\textit{AutoGraph-R1 demonstrably improves the graph's core function as a knowledge index.}} The magnitude of F1 gains is more modest in the text retrieval setting. This is an expected outcome, stemming from the dual nature of using full text passages as evidence. On one hand, their rich context can enable the generator to succeed even with imperfect retrieval, masking some F1 gains. On the other hand, this verbosity can introduce noise, unlike the concise, structured triples provided by the graph knowledge retriever. To isolate the direct impact on retrieval quality, we evaluate passage recall@5. Table \ref{tab:unified_results} reveals significant improvements across all models. We observe consistent gains in average recall (e.g., + over 4.0 points for Qwen-3B), confirming that our RL framework successfully optimizes the graph structure to act as a more effective index, reliably guiding text retrievers to the correct evidence.

\begin{table*}[!ht]
\vspace{-0.1in}
\small
\centering
\caption{Further analysis on whether GRPO training increases the triple extraction intrinsic quality, measured by Precision, Recall and, F1 defined in previous work \citep{huang_can_2025,bai_autoschemakg_2025}.}
\resizebox{\textwidth}{!}{
\begin{NiceTabular}{@{}l|ccc|ccc|ccc|ccc@{}}
\toprule
\multirow{2}{*}{KG Construction Model} & \multicolumn{3}{c}{HotpotQA} & \multicolumn{3}{c}{2WikiMultihopQA} & \multicolumn{3}{c}{Musique} & \multicolumn{3}{c}{2021Wiki} \\
\cmidrule{2-13}
 & Acc & Recall & F1 & Acc & Recall & F1 & Acc & Recall & F1 & Acc & Recall & F1 \\
\midrule
Qwen2.5-7B-Instruct  & 98.50 & 93.68 & 95.65 & 94.80 & 91.19 & 92.68 & 96.77 & 95.27 & 95.73 & 95.03 & 91.39 & 92.92 \\
$\thickspace$ + GRPO with Knowledge-Carrying Reward & 97.53 & \textbf{96.66} &\textbf{ 96.71} & 95.51 & \textbf{96.55} & 95.25 & 97.14 & \textbf{96.75} & \textbf{96.45} & 96.17 & \textbf{96.66} & \textbf{96.15} \\
$\thickspace$ + GRPO with Knowledge-Indexing Reward & \textbf{98.96}& 94.81 & 96.59 & \textbf{98.35} & 94.54 & \textbf{96.16} & \textbf{99.53} & 93.14 & 95.81 & \textbf{97.44} & 95.01 & 95.99 \\
\midrule
Qwen2.5-3B-Instruct  & 94.41 & 91.00 & 91.92 & 83.53 & 79.34 & 81.01 & 92.07 & 89.52 & 90.31 & 87.79 & 86.01 & 86.63 \\
 $\thickspace$ + GRPO with Knowledge-Carrying Reward  & 96.52 & \textbf{94.24} & 94.80 & 95.91 & \textbf{96.28} &\textbf{ 95.80} & \textbf{97.01 }& \textbf{94.74} & \textbf{95.22} & 96.70 & 95.58 & 95.76 \\
 $\thickspace$ + GRPO with Knowledge-Indexing Reward  & \textbf{97.11} & 93.15 & \textbf{94.64} & \textbf{96.19} & 93.66 & 94.48 & 96.20 & 93.87 & 94.55 & \textbf{98.22 }& \textbf{96.04 }& \textbf{96.85} \\
\bottomrule
\bottomrule
\end{NiceTabular}}
\label{tab:performance_metrics}
\vspace{-0.2in}
\end{table*}

\textbf{\textit{Optimizing for downstream utility also enhances the intrinsic factual quality of the graph.}} We investigated whether extrinsic optimization comes at the cost of intrinsic quality by measuring the precision, recall, and F1 score of the extracted triples against the source text \citep{huang_can_2025} using Deepseek-V3 model as a judge \citep{deepseek-ai_deepseek-v3_2025}. The results in Table \ref{tab:performance_metrics} show a clear positive correlation. Across all datasets, KGs fine-tuned with AutoGraph-R1 exhibit higher intrinsic F1 scores than their zero-shot counterparts. This indicates our RL framework does not sacrifice factual accuracy for functional utility; rather, it improves both simultaneously.

\textbf{\textit{The choice of reward function induces specific and beneficial structural biases in the KG.}} A deeper analysis of Table \ref{tab:performance_metrics} reveals that the two reward functions specialize the graph's structure. The \textbf{Knowledge-Carrying Reward} ($R_C$), optimized for graph knowledge retrieval, consistently produces graphs with higher recall, aligning with its goal of ensuring all necessary facts for reasoning are present. In contrast, the \textbf{Knowledge-Indexing Reward} ($R_I$), optimized for text retrieval, yields graphs with higher precision, reflecting its need for a clean, high-fidelity index. This finding highlights that AutoGraph-R1 not only improves graph quality but also tailors the graph's structure to its specific downstream function. For analysis on comparisons with 70B+ baselines, On-Policy RL versus offline fine-tuning, and the structural analysis of fintuned LLM constructed KG, please refer to the ablation studies in Appendix \ref{sec:appendix_ablation}.
\subsection{Ablation: Impact of Reward Function Design}
To validate our task-specific rewards ($R_C$ and $R_I$), we compare them against the final answer's F1 score as a direct reward signal. We trained two additional KG constructor models using this F1 reward, keeping all other hyperparameters identical to our main setup. As shown in 
Table~\ref{tab:ablation_unified}, the F1-rewarded model underperforms the zero-shot 
baseline by over $-$2.2 average F1 on the Triples Retriever and actively degrades 
both HippoRAG and HippoRAG2 performance. In contrast, our task-specific rewards 
deliver consistent gains across all settings, confirming that measuring functional 
graph utility directly is essential for stable RL-based KG construction. Detailed 
training dynamics are provided in Appendix~\ref{sec:f1_reward_autograph-r1}.

\begin{table*}[!ht]
\small
\centering
\caption{Ablation Study on Reward Functions for Qwen2.5-7B. \textbf{(Top)} Graph Knowledge Retrievers evaluated on F1. \textbf{(Bottom)} Graph-based Text Retrievers evaluated on F1 and Recall@5. Our task-specific rewards ($R_C$, $R_I$) consistently outperform the naive F1 Reward and zero-shot baseline across settings.}
\label{tab:ablation_unified}
\resizebox{0.88\textwidth}{!}{
\begin{NiceTabular}{@{}l|cc|cc|cc|cc|cc|cc@{}}
\toprule
\multirow{2}{*}{\textbf{Methods}}
  & \multicolumn{2}{c|}{\textbf{NQ*}}
  & \multicolumn{2}{c|}{\textbf{PopQA*}}
  & \multicolumn{2}{c|}{\textbf{HotpotQA}}
  & \multicolumn{2}{c|}{\textbf{2WikiMultihopQA}}
  & \multicolumn{2}{c|}{\textbf{Musique}}
  & \multicolumn{2}{c}{\textbf{Avg.}} \\
& \multicolumn{2}{c|}{F1}
& \multicolumn{2}{c|}{F1}
& \multicolumn{2}{c|}{F1}
& \multicolumn{2}{c|}{F1}
& \multicolumn{2}{c|}{F1}
& \multicolumn{2}{c}{F1} \\
\midrule

\multicolumn{13}{c}{\cellcolor[gray]{0.9}\textbf{Graph Knowledge Retrievers (F1 Only)}} \\
\midrule
Subgraph (Base)
  & \multicolumn{2}{c|}{28.07}
  & \multicolumn{2}{c|}{55.43}
  & \multicolumn{2}{c|}{41.66}
  & \multicolumn{2}{c|}{33.97}
  & \multicolumn{2}{c|}{15.24}
  & \multicolumn{2}{c}{34.87} \\
Subgraph (F1 Reward)
  & \multicolumn{2}{c|}{27.36\textcolor{blue}{$\downarrow_{0.71}$}}
  & \multicolumn{2}{c|}{55.09\textcolor{blue}{$\downarrow_{0.34}$}}
  & \multicolumn{2}{c|}{41.15\textcolor{blue}{$\downarrow_{0.51}$}}
  & \multicolumn{2}{c|}{35.13}
  & \multicolumn{2}{c|}{15.70}
  & \multicolumn{2}{c}{34.89} \\
\rowcolor{beaublue}
Subgraph ($R_C$)
  & \multicolumn{2}{c|}{\textbf{28.54}}
  & \multicolumn{2}{c|}{\textbf{60.94}}
  & \multicolumn{2}{c|}{\textbf{43.59}}
  & \multicolumn{2}{c|}{\textbf{37.43}}
  & \multicolumn{2}{c|}{\textbf{15.65}}
  & \multicolumn{2}{c}{\textbf{37.23}} \\
\midrule
Triples (Base)
  & \multicolumn{2}{c|}{33.26}
  & \multicolumn{2}{c|}{55.56}
  & \multicolumn{2}{c|}{44.99}
  & \multicolumn{2}{c|}{35.57}
  & \multicolumn{2}{c|}{20.43}
  & \multicolumn{2}{c}{37.96} \\
Triples (F1 Reward)
  & \multicolumn{2}{c|}{31.52\textcolor{blue}{$\downarrow_{1.74}$}}
  & \multicolumn{2}{c|}{53.85\textcolor{blue}{$\downarrow_{1.71}$}}
  & \multicolumn{2}{c|}{44.52\textcolor{blue}{$\downarrow_{0.47}$}}
  & \multicolumn{2}{c|}{30.68\textcolor{blue}{$\downarrow_{4.89}$}}
  & \multicolumn{2}{c|}{18.00\textcolor{blue}{$\downarrow_{2.43}$}}
  & \multicolumn{2}{c}{35.71\textcolor{blue}{$\downarrow_{2.25}$}} \\
\rowcolor{beaublue}
Triples ($R_C$)
  & \multicolumn{2}{c|}{\textbf{33.98}}
  & \multicolumn{2}{c|}{\textbf{58.02}}
  & \multicolumn{2}{c|}{\textbf{48.28}}
  & \multicolumn{2}{c|}{\textbf{36.04}}
  & \multicolumn{2}{c|}{\textbf{20.56}}
  & \multicolumn{2}{c}{\textbf{39.38}} \\
\midrule
ToG (Base)
  & \multicolumn{2}{c|}{25.59}
  & \multicolumn{2}{c|}{57.53}
  & \multicolumn{2}{c|}{43.93}
  & \multicolumn{2}{c|}{46.03}
  & \multicolumn{2}{c|}{18.46}
  & \multicolumn{2}{c}{38.31} \\
ToG (F1 Reward)
  & \multicolumn{2}{c|}{27.64}
  & \multicolumn{2}{c|}{56.95\textcolor{blue}{$\downarrow_{0.58}$}}
  & \multicolumn{2}{c|}{45.19}
  & \multicolumn{2}{c|}{51.10}
  & \multicolumn{2}{c|}{18.37\textcolor{blue}{$\downarrow_{0.09}$}}
  & \multicolumn{2}{c}{39.85} \\
\rowcolor{beaublue}
ToG ($R_C$)
  & \multicolumn{2}{c|}{\textbf{29.36}}
  & \multicolumn{2}{c|}{\textbf{62.85}}
  & \multicolumn{2}{c|}{\textbf{44.68}}
  & \multicolumn{2}{c|}{\textbf{50.20}}
  & \multicolumn{2}{c|}{\textbf{19.31}}
  & \multicolumn{2}{c}{\textbf{41.28}} \\
\midrule\midrule

\multicolumn{13}{c}{\cellcolor[gray]{0.9}\textbf{Graph-based Text Retrievers (F1 \& Recall@5)}} \\
\midrule
& F1 & R@5 & F1 & R@5 & F1 & R@5 & F1 & R@5 & F1 & R@5 & F1 & R@5 \\
\midrule
HippoRAG (Base)
  & 37.16 & 93.30 & 65.95 & 92.90 & 55.50 & 69.97 & 53.01 & 72.87 & 26.03 & 47.19 & 47.53 & 75.25 \\
HippoRAG (F1 Reward)
  & 39.66 & 93.90 & 63.74\textcolor{blue}{$\downarrow_{2.21}$} & 92.10\textcolor{blue}{$\downarrow_{0.80}$} & 53.74\textcolor{blue}{$\downarrow_{1.76}$} & 69.18\textcolor{blue}{$\downarrow_{0.79}$} & 49.58\textcolor{blue}{$\downarrow_{3.43}$} & 72.71\textcolor{blue}{$\downarrow_{0.16}$} & 28.68 & 47.71 & 47.08\textcolor{blue}{$\downarrow_{0.45}$} & 75.12\textcolor{blue}{$\downarrow_{0.13}$} \\
\rowcolor{beaublue}
HippoRAG ($R_I$)
  & \textbf{38.80} & \textbf{94.30} & \textbf{67.85} & \textbf{95.80} & \textbf{57.19} & \textbf{71.40} & \textbf{53.60} & \textbf{76.16} & \textbf{26.97} & \textbf{48.44} & \textbf{48.88} & \textbf{77.22} \\
\midrule
HippoRAG2 (Base)
  & 37.02 & 94.10 & 65.74 & 92.80 & 57.08 & 72.03 & 54.99 & 75.98 & 26.77 & 48.55 & 48.32 & 76.69 \\
HippoRAG2 (F1 Reward)
  & 38.33 & 94.80 & 62.92\textcolor{blue}{$\downarrow_{2.82}$} & 92.00\textcolor{blue}{$\downarrow_{0.80}$} & 55.19\textcolor{blue}{$\downarrow_{1.89}$} & 71.07\textcolor{blue}{$\downarrow_{0.96}$} & 50.23\textcolor{blue}{$\downarrow_{4.76}$} & 72.81\textcolor{blue}{$\downarrow_{3.17}$} & 27.51 & 48.75 & 46.83\textcolor{blue}{$\downarrow_{1.49}$} & 75.88\textcolor{blue}{$\downarrow_{0.81}$} \\
\rowcolor{beaublue}
HippoRAG2 ($R_I$)
  & \textbf{38.68} & \textbf{95.00} & \textbf{67.72} & \textbf{96.30} & \textbf{58.98} & \textbf{73.61} & \textbf{56.46} & \textbf{78.66} & \textbf{27.18} & \textbf{49.23} & \textbf{49.80} & \textbf{78.56} \\
\bottomrule
\end{NiceTabular}}
\vspace{-0.1in}
\end{table*}

\subsection{Computational Cost}
\label{sec:computational_cost}
The per-sample training cost is bounded: graph construction operates over a small, 
fixed document pool (15 documents for text retrieval; 2--5 for graph retrieval), and 
the $R_C$ judge prompt (Figure~\ref{fig:prompt_deducibility_judge}) is a minimal binary Yes/No query that can be batched 
efficiently. Table~\ref{tab:training_time} reports wall-clock times on 2$\times$H100 
GPUs, which are well within academic lab budgets.

\begin{table}[!ht]
\centering
\small
\caption{Wall-clock training times for AutoGraph-R1 on 2$\times$H100 GPUs.}
\begin{tabular}{lc}
\toprule
\textbf{Experiment Setting} & \textbf{Time} \\
\midrule
7B Graph-based Text Retriever & $\sim$8h 38m \\
3B Graph-based Text Retriever & $\sim$5h 11m \\
7B Graph Knowledge Retriever  & $\sim$3h 15m \\
3B Graph Knowledge Retriever  & $\sim$2h 18m \\
\bottomrule
\end{tabular}
\label{tab:training_time}
\end{table}

\section{Conclusion}
In this work, we introduced \textbf{AutoGraph-R1}, the first reinforcement learning framework for knowledge graph construction that directly optimizes downstream RAG performance. By incorporating task-aware rewards, our approach bridges the gap between traditional graph quality metrics and end-task utility. Experiments across five QA benchmarks demonstrate consistent improvements over strong baselines in both graph knowledge and graph-based text retrieval. Overall, our work shows that reinforcement learning can effectively connect the graph construction process with downstream QA performance, ensuring that knowledge graphs are optimized for their intended applications.

\section{Limitations}
\label{sec:limitations}

While AutoGraph-R1 advances the state of task-aware knowledge graph construction, several limitations warrant discussion:

\textbf{Dependency on Frozen Retrievers.} Our framework optimizes the graph construction policy to maximize utility for a \textit{specific, frozen} retrieval module. While this ensures high performance for the target pipeline, it implicitly couples the graph structure to the retriever's inductive biases. A graph optimized for a graph knowledge retriever effectively learns to construct reasoning pathways, which may not be the optimal topology for a distinct retrieval paradigm (e.g., a BM25 retriever).

\textbf{Reward Computation Bottleneck.} Our use of task-aware rewards ($R_C$ and $R_I$) necessitates executing a full retrieval and evaluation loop during training. For the Knowledge-Carrying reward specifically, this involves querying a LLM judge for every generated sample to verify deducibility. This introduces computational overhead compared to offline or heuristic-based training objectives, potentially limiting the scalability of our method to extremely large training datasets without substantial compute resources.

\textbf{Model Scale and Generalization.} Due to computational constraints, our experiments were primarily conducted on LLMs in the 1B to 7B parameter range (Qwen2.5 and Llama-3.2). While we demonstrate that fine-tuned smaller models can outperform larger zero-shot baselines, we have not yet empirically verified the scaling laws of our RL framework on massive foundation models (e.g., 70B+ parameters). Additionally, our evaluation is limited to English-language benchmarks; the effectiveness of our reward-driven policy in multilingual or low-resource settings remains an open question.

\textbf{Dependency on Associated Passages.} Our training pipeline assumes access to associated passages alongside query-answer pairs, i.e., training data of the form \textit{(query, gold\_answer, associated\_passages)}. While such data is available in standard multi-hop QA benchmarks (e.g., HotpotQA, Musique), this assumption may not hold in more general settings where only \textit{(query, gold\_answer)} pairs are available. Collecting associated passages constitutes an additional annotation cost, which may limit the applicability of our framework in low-resource or domain-specific scenarios. We leave the exploration of passage-free training strategies as an important direction for future work.

\section{Ethics Statement}
This work complies with the ACL Code of Ethics. Our research focuses on optimizing the structural utility of knowledge graphs and does not involve human subjects or the collection of non-public data. All experiments utilize established, publicly available datasets (HotpotQA, Musique, NQ, PopQA). We acknowledge that generative models and retrieval-augmented systems can inadvertently propagate biases present in their training corpora (e.g., Wikipedia) or pre-trained base models. Furthermore, while our method aims to improve retrieval accuracy, the automated construction of knowledge graphs carries a risk of generating or amplifying hallucinations if the source text is factually incorrect. We encourage practitioners to audit both the source corpora and the constructed graphs, particularly when applying this framework in sensitive domains such as healthcare or finance. This paper uses LLMs to assist with grammar and writing quality.

\section*{Acknowledgments}
The authors of this paper were supported by the ITSP Platform Research Project (ITS/189/23FP) from ITC of Hong Kong, SAR, China, and the AoE (AoE/E-601/24-N), the RIF (R6021-20), the GRF (16205322) and the JRFS (JRFS2526-6S10) from RGC of Hong Kong, SAR, China. We also thank the support from the Microsoft Accelerate Foundation Models Research (AFMR) grant program.
\bibliography{custom}

\clearpage
\appendix

\section{Ablation Study}
\label{sec:appendix_ablation}
\subsection{Comparison with State-of-the-Art Large Models}
To assess the efficacy of our reward-driven optimization, we compared our fine-tuned 3B and 7B models against a suite of state-of-the-art (SOTA) large-scale models, including Llama-3.3-70B, Qwen2.5-72B, and GPT-5-mini and etc.

Table \ref{tab:unified_sota_results} present the results. Our findings demonstrate that AutoGraph-R1 allows smaller models to achieve parity with, and often surpass, models that are significantly larger. Specifically: \begin{itemize} \item On Graph Retrievers, our Qwen-7B model demonstrates remarkable efficiency, surpassing most zero-shot models on both Subgraph and Triples retriever, while maintaining competitive parity on the ToG retriever. \item On Text Retrievers, our Qwen-7B model outperforms the baselines for HippoRAG and HippoRAG2, validating that our method builds more effective knowledge indices than general-purpose large models. \end{itemize}

\subsection{Ablation Study on Training Methodology (RL vs. RAFT)}
To determine if our performance gains stem specifically from On-Policy RL (GRPO) or simply from reward-guided data selection, we conducted an ablation study comparing GRPO against a RAFT (Reward rAnked FineTuning) approach \citep{dong_raft_2023}. In the RAFT setup, we utilized our reward functions to filter generated KGs offline and fine-tuned the model on high-reward samples.

Table \ref{tab:raft_ablation} presents the results on the Qwen2.5-7B model. Both strategies significantly outperform the zero-shot baseline, confirming that our reward functions capture genuine signal for downstream utility. Comparing the two, GRPO consistently yields superior results, achieving the highest average F1 score in 4 out of 5 retrieval settings. While RAFT remains a highly competitive and computationally efficient alternative, even slightly outperforming GRPO on the ToG retriever the results suggest that the active exploration inherent in on-policy RL enables the model to discover more robust structural strategies than offline filtering alone.

\begin{table*}[!p]
\small
\centering
\caption{Comprehensive performance evaluation of various large-scale models as KG constructors. The table compares zero-shot KG construction performance across Graph Knowledge Retrievers (F1 Only) and Graph-based Text Retrievers (F1 \& Recall). Results show that even powerful models like GPT-5-mini and Llama-3.3-70B exhibit performance variances, highlighting the importance of task-specific optimization.}
\label{tab:unified_sota_results}
\resizebox{0.8\textwidth}{!}{
\begin{NiceTabular}{@{}l|cc|cc|cc|cc|cc|cc@{}}
\toprule
\multirow{2}{*}{\textbf{Methods}}
  & \multicolumn{2}{c|}{\textbf{NQ*}}
  & \multicolumn{2}{c|}{\textbf{PopQA*}}
  & \multicolumn{2}{c|}{\textbf{HotpotQA}}
  & \multicolumn{2}{c|}{\textbf{2WikiMultihopQA}}
  & \multicolumn{2}{c|}{\textbf{Musique}}
  & \multicolumn{2}{c}{\textbf{Avg.}} \\
& \multicolumn{2}{c|}{F1}
& \multicolumn{2}{c|}{F1}
& \multicolumn{2}{c|}{F1}
& \multicolumn{2}{c|}{F1}
& \multicolumn{2}{c|}{F1}
& \multicolumn{2}{c}{F1} \\
\midrule

\multicolumn{13}{c}{\cellcolor[gray]{0.9}\textbf{Graph Knowledge Retrievers (F1 Only)}} \\
\midrule

\rowcolor[gray]{0.95}{\textit{DeepSeek-V3}} \\
\midrule
Subgraph Retriever & \multicolumn{2}{c|}{27.20} & \multicolumn{2}{c|}{61.42} & \multicolumn{2}{c|}{42.36} & \multicolumn{2}{c|}{34.96} & \multicolumn{2}{c|}{13.29} & \multicolumn{2}{c}{35.85} \\
Triples Retriever  & \multicolumn{2}{c|}{33.89} & \multicolumn{2}{c|}{54.61} & \multicolumn{2}{c|}{44.84} & \multicolumn{2}{c|}{32.67} & \multicolumn{2}{c|}{19.13} & \multicolumn{2}{c}{37.03} \\
ToG Retriever      & \multicolumn{2}{c|}{28.29} & \multicolumn{2}{c|}{64.02} & \multicolumn{2}{c|}{44.72} & \multicolumn{2}{c|}{50.98} & \multicolumn{2}{c|}{17.99} & \multicolumn{2}{c}{41.20} \\
\midrule

\rowcolor[gray]{0.95}{\textit{GPT-5-mini}} \\
\midrule
Subgraph Retriever & \multicolumn{2}{c|}{27.09} & \multicolumn{2}{c|}{58.62} & \multicolumn{2}{c|}{41.25} & \multicolumn{2}{c|}{32.78} & \multicolumn{2}{c|}{14.34} & \multicolumn{2}{c}{34.82} \\
Triples Retriever  & \multicolumn{2}{c|}{34.62} & \multicolumn{2}{c|}{54.99} & \multicolumn{2}{c|}{43.07} & \multicolumn{2}{c|}{29.66} & \multicolumn{2}{c|}{18.29} & \multicolumn{2}{c}{36.13} \\
ToG Retriever      & \multicolumn{2}{c|}{28.20} & \multicolumn{2}{c|}{62.65} & \multicolumn{2}{c|}{44.14} & \multicolumn{2}{c|}{49.11} & \multicolumn{2}{c|}{18.78} & \multicolumn{2}{c}{40.58} \\
\midrule

\rowcolor[gray]{0.95}{\textit{Llama-3.3-70B}} \\
\midrule
Subgraph Retriever & \multicolumn{2}{c|}{26.90} & \multicolumn{2}{c|}{62.05} & \multicolumn{2}{c|}{41.81} & \multicolumn{2}{c|}{34.56} & \multicolumn{2}{c|}{16.11} & \multicolumn{2}{c}{36.29} \\
Triples Retriever  & \multicolumn{2}{c|}{32.95} & \multicolumn{2}{c|}{54.65} & \multicolumn{2}{c|}{44.00} & \multicolumn{2}{c|}{31.84} & \multicolumn{2}{c|}{19.55} & \multicolumn{2}{c}{36.60} \\
ToG Retriever      & \multicolumn{2}{c|}{26.31} & \multicolumn{2}{c|}{62.48} & \multicolumn{2}{c|}{45.45} & \multicolumn{2}{c|}{50.42} & \multicolumn{2}{c|}{18.59} & \multicolumn{2}{c}{40.65} \\
\midrule

\rowcolor[gray]{0.95}{\textit{gpt-oss-120b}} \\
\midrule
Subgraph Retriever & \multicolumn{2}{c|}{26.92} & \multicolumn{2}{c|}{58.83} & \multicolumn{2}{c|}{40.60} & \multicolumn{2}{c|}{33.35} & \multicolumn{2}{c|}{13.39} & \multicolumn{2}{c}{34.62} \\
Triples Retriever  & \multicolumn{2}{c|}{32.95} & \multicolumn{2}{c|}{54.43} & \multicolumn{2}{c|}{42.91} & \multicolumn{2}{c|}{29.66} & \multicolumn{2}{c|}{19.10} & \multicolumn{2}{c}{35.81} \\
ToG Retriever      & \multicolumn{2}{c|}{26.00} & \multicolumn{2}{c|}{60.82} & \multicolumn{2}{c|}{43.52} & \multicolumn{2}{c|}{48.43} & \multicolumn{2}{c|}{18.30} & \multicolumn{2}{c}{39.41} \\
\midrule

\rowcolor[gray]{0.95}{\textit{Qwen2.5-72B-Instruct}} \\
\midrule
Subgraph Retriever & \multicolumn{2}{c|}{28.12} & \multicolumn{2}{c|}{62.23} & \multicolumn{2}{c|}{42.57} & \multicolumn{2}{c|}{36.17} & \multicolumn{2}{c|}{14.85} & \multicolumn{2}{c}{36.79} \\
Triples Retriever  & \multicolumn{2}{c|}{34.23} & \multicolumn{2}{c|}{54.68} & \multicolumn{2}{c|}{44.10} & \multicolumn{2}{c|}{32.47} & \multicolumn{2}{c|}{20.07} & \multicolumn{2}{c}{37.11} \\
ToG Retriever      & \multicolumn{2}{c|}{28.73} & \multicolumn{2}{c|}{64.21} & \multicolumn{2}{c|}{45.54} & \multicolumn{2}{c|}{50.19} & \multicolumn{2}{c|}{18.37} & \multicolumn{2}{c}{41.41} \\
\midrule

\rowcolor[gray]{0.95}{\textit{Qwen3-235B}} \\
\midrule
Subgraph Retriever & \multicolumn{2}{c|}{27.62} & \multicolumn{2}{c|}{60.90} & \multicolumn{2}{c|}{41.49} & \multicolumn{2}{c|}{35.59} & \multicolumn{2}{c|}{15.83} & \multicolumn{2}{c}{36.29} \\
Triples Retriever  & \multicolumn{2}{c|}{34.12} & \multicolumn{2}{c|}{55.12} & \multicolumn{2}{c|}{41.67} & \multicolumn{2}{c|}{32.81} & \multicolumn{2}{c|}{20.71} & \multicolumn{2}{c}{36.89} \\
ToG Retriever      & \multicolumn{2}{c|}{27.68} & \multicolumn{2}{c|}{63.54} & \multicolumn{2}{c|}{45.35} & \multicolumn{2}{c|}{49.68} & \multicolumn{2}{c|}{21.77} & \multicolumn{2}{c}{41.60} \\
\midrule

\rowcolor[gray]{0.95}{\textit{gemini-2.5-flash-lite}} \\
\midrule
Subgraph Retriever & \multicolumn{2}{c|}{27.81} & \multicolumn{2}{c|}{60.42} & \multicolumn{2}{c|}{40.95} & \multicolumn{2}{c|}{35.21} & \multicolumn{2}{c|}{15.18} & \multicolumn{2}{c}{35.91} \\
Triples Retriever  & \multicolumn{2}{c|}{33.52} & \multicolumn{2}{c|}{54.73} & \multicolumn{2}{c|}{44.02} & \multicolumn{2}{c|}{33.02} & \multicolumn{2}{c|}{19.91} & \multicolumn{2}{c}{37.04} \\
ToG Retriever      & \multicolumn{2}{c|}{25.90} & \multicolumn{2}{c|}{62.55} & \multicolumn{2}{c|}{42.69} & \multicolumn{2}{c|}{49.55} & \multicolumn{2}{c|}{19.60} & \multicolumn{2}{c}{40.06} \\
\midrule

\rowcolor{beaublue}{\textbf{AutoGraph-R1 (Qwen-3B)}} \\
\midrule
Subgraph Retriever & \multicolumn{2}{c|}{28.03} & \multicolumn{2}{c|}{59.46} & \multicolumn{2}{c|}{40.77} & \multicolumn{2}{c|}{34.71} & \multicolumn{2}{c|}{15.13} & \multicolumn{2}{c}{35.62} \\
Triples Retriever  & \multicolumn{2}{c|}{33.67} & \multicolumn{2}{c|}{56.76} & \multicolumn{2}{c|}{46.94} & \multicolumn{2}{c|}{36.09} & \multicolumn{2}{c|}{21.41} & \multicolumn{2}{c}{38.97} \\
ToG Retriever      & \multicolumn{2}{c|}{29.27} & \multicolumn{2}{c|}{61.40} & \multicolumn{2}{c|}{44.56} & \multicolumn{2}{c|}{49.33} & \multicolumn{2}{c|}{18.42} & \multicolumn{2}{c}{40.60} \\
\midrule

\rowcolor{beaublue}{\textbf{AutoGraph-R1 (Qwen-7B)}} \\
\midrule
Subgraph Retriever & \multicolumn{2}{c|}{28.54} & \multicolumn{2}{c|}{60.94} & \multicolumn{2}{c|}{43.59} & \multicolumn{2}{c|}{37.43} & \multicolumn{2}{c|}{15.65} & \multicolumn{2}{c}{37.23} \\
Triples Retriever  & \multicolumn{2}{c|}{33.98} & \multicolumn{2}{c|}{58.02} & \multicolumn{2}{c|}{48.28} & \multicolumn{2}{c|}{36.04} & \multicolumn{2}{c|}{20.56} & \multicolumn{2}{c}{39.38} \\
ToG Retriever      & \multicolumn{2}{c|}{29.36} & \multicolumn{2}{c|}{62.85} & \multicolumn{2}{c|}{44.68} & \multicolumn{2}{c|}{50.20} & \multicolumn{2}{c|}{19.31} & \multicolumn{2}{c}{41.28} \\
\midrule\midrule\midrule

\multicolumn{13}{c}{\cellcolor[gray]{0.9}\textbf{Graph-based Text Retrievers (F1 \& Recall@5)}} \\
\midrule
& F1 & R@5 & F1 & R@5 & F1 & R@5 & F1 & R@5 & F1 & R@5 & F1 & R@5 \\
\midrule

\rowcolor[gray]{0.95}{\textit{DeepSeek-V3}} \\
\midrule
HippoRAG  & 39.82 & 93.90 & 66.29 & 94.90 & 54.33 & 68.30 & 49.59 & 72.69 & 27.79 & 47.23 & 47.56 & 75.40 \\
HippoRAG2 & 38.45 & 93.70 & 65.73 & 96.10 & 54.89 & 69.68 & 52.23 & 75.17 & 28.14 & 47.76 & 47.89 & 76.48 \\
\midrule

\rowcolor[gray]{0.95}{\textit{GPT-5-mini}} \\
\midrule
HippoRAG  & 39.28 & 93.80 & 66.12 & 94.70 & 54.22 & 68.60 & 47.59 & 69.92 & 28.70 & 47.80 & 47.18 & 74.96 \\
HippoRAG2 & 39.20 & 94.20 & 65.47 & 94.20 & 55.85 & 69.28 & 49.44 & 69.77 & 28.80 & 49.00 & 47.75 & 75.29 \\
\midrule

\rowcolor[gray]{0.95}{\textit{Llama-3.3-70B}} \\
\midrule
HippoRAG  & 39.29 & 94.30 & 63.57 & 94.00 & 56.14 & 60.00 & 52.70 & 73.12 & 28.32 & 47.41 & 48.00 & 73.77 \\
HippoRAG2 & 39.29 & 94.50 & 65.18 & 95.50 & 56.29 & 70.71 & 53.93 & 75.89 & 26.97 & 48.27 & 48.33 & 76.97 \\
\midrule

\rowcolor[gray]{0.95}{\textit{gpt-oss-120b}} \\
\midrule
HippoRAG  & 39.52 & 93.70 & 64.82 & 92.90 & 53.11 & 65.61 & 47.79 & 68.57 & 25.37 & 45.43 & 46.12 & 73.24 \\
HippoRAG2 & 39.45 & 93.60 & 64.78 & 93.10 & 54.53 & 66.51 & 49.49 & 69.49 & 27.03 & 46.97 & 47.06 & 73.93 \\
\midrule

\rowcolor[gray]{0.95}{\textit{Qwen2.5-72B-Instruct}} \\
\midrule
HippoRAG  & 39.73 & 94.60 & 65.18 & 94.00 & 53.98 & 67.92 & 50.68 & 73.06 & 26.42 & 46.93 & 47.20 & 75.30 \\
HippoRAG2 & 39.29 & 95.10 & 65.71 & 94.60 & 57.34 & 70.42 & 53.21 & 76.00 & 26.71 & 48.63 & 48.45 & 76.95 \\
\midrule

\rowcolor[gray]{0.95}{\textit{Qwen3-235B}} \\
\midrule
HippoRAG  & 39.56 & 93.90 & 65.32 & 95.40 & 55.07 & 68.02 & 51.94 & 72.67 & 27.72 & 47.58 & 47.92 & 75.51 \\
HippoRAG2 & 39.68 & 94.30 & 67.25 & 95.60 & 54.81 & 68.52 & 51.76 & 74.37 & 28.91 & 48.13 & 48.48 & 76.18 \\
\midrule

\rowcolor[gray]{0.95}{\textit{gemini-2.5-flash-lite}} \\
\midrule
HippoRAG  & 40.06 & 94.40 & 64.54 & 94.40 & 55.39 & 66.88 & 51.55 & 72.19 & 27.32 & 46.36 & 47.77 & 74.85 \\
HippoRAG2 & 39.73 & 94.10 & 65.48 & 95.30 & 54.55 & 68.93 & 51.73 & 74.37 & 28.34 & 47.81 & 47.97 & 76.10 \\
\midrule

\rowcolor{beaublue}{\textbf{AutoGraph-R1 (Qwen-3B)}} \\
\midrule
HippoRAG  & 38.28 & 93.00 & 65.93 & 95.60 & 55.39 & 68.82 & 51.69 & 74.02 & 28.11 & 47.65 & 47.88 & 75.82 \\
HippoRAG2 & 38.45 & 94.00 & 66.23 & 95.40 & 56.28 & 71.21 & 52.80 & 76.42 & 27.93 & 49.13 & 48.34 & 77.23 \\
\midrule

\rowcolor{beaublue}{\textbf{AutoGraph-R1 (Qwen-7B)}} \\
\midrule
HippoRAG  & 38.80 & 94.30 & 67.85 & 95.80 & 57.19 & 71.40 & 53.60 & 76.16 & 26.97 & 48.44 & 48.88 & 77.22 \\
HippoRAG2 & 38.68 & 95.00 & 67.72 & 96.30 & 58.98 & 73.61 & 56.46 & 78.66 & 27.18 & 49.23 & 49.80 & 78.56 \\
\bottomrule
\end{NiceTabular}}
\end{table*}

\begin{table*}[p]
\centering
\small
\caption{Ablation study comparing On-Policy RL (GRPO) against Offline Filtering (RAFT) for Qwen2.5-7B. Both methods leverage our reward functions. Results show that reward-guided training consistently improves over the baseline, with GRPO and RAFT offering competitive trade-offs.}
\resizebox{0.8\textwidth}{!}{
\begin{NiceTabular}{@{}l|l|llllll@{}}
\toprule
\textbf{Retriever} & \textbf{Method} & \textbf{NQ} & \textbf{PopQA} & \textbf{HotpotQA} & \textbf{2Wiki} & \textbf{Musique} & \textbf{Avg.} \\
\midrule
\multicolumn{8}{c}{\cellcolor[gray]{0.9}\textit{Graph Retrievers (Knowledge-Carrying Reward)}} \\
\midrule
\multirow{3}{*}{Subgraph} & Base & 28.07 & 55.43 & 41.66 & 33.97 & 15.24 & 34.87 \\
 & GRPO (On-Policy) & 28.54 & 60.94 & \textbf{43.59} & \textbf{37.43} & \textbf{15.65} & \textbf{37.23} \\
 & RAFT (Offline) & \textbf{28.72} & \textbf{61.27} & 42.14 & 35.78 & 15.41 & 36.66 \\
\midrule
\multirow{3}{*}{Triples} & Base & 33.26 & 55.56 & 44.99 & 35.57 & 20.43 & 37.96 \\
 & GRPO (On-Policy) & 33.98 & \textbf{58.02} & \textbf{48.28} & \textbf{36.04} & \textbf{20.56} & \textbf{39.38} \\
 & RAFT (Offline) & \textbf{34.72} & 56.42 & 44.91 & 31.78 & 20.21 & 37.61 \\
\midrule
\multirow{3}{*}{ToG} & Base & 25.59 & 57.53 & 43.93 & 46.03 & 18.46 & 38.31 \\
 & GRPO (On-Policy) & 29.36 & 62.85 & 44.68 & \textbf{50.20} & 19.31 & 41.28 \\
 & RAFT (Offline) & \textbf{29.65} & \textbf{63.30} & \textbf{45.47} & 50.10 & \textbf{20.32} & \textbf{41.77} \\
\midrule
\multicolumn{8}{c}{\cellcolor[gray]{0.9}\textit{Text Retrievers (Knowledge-Indexing Reward)}} \\
\midrule
\multirow{3}{*}{HippoRAG} & Base & 37.16 & 65.95 & 55.50 & 53.01 & 26.03 & 47.53 \\
 & GRPO (On-Policy) & 38.80 & \textbf{67.85} & \textbf{57.19} & \textbf{53.60} & 26.97 & \textbf{48.88} \\
 & RAFT (Offline) & \textbf{39.60} & 66.22 & 55.77 & 51.01 & \textbf{29.29} & 48.38 \\
\midrule
\multirow{3}{*}{HippoRAG2} & Base & 37.02 & 65.74 & 57.08 & 54.99 & 26.77 & 48.32 \\
 & GRPO (On-Policy) & 38.68 & \textbf{67.72} & \textbf{58.98} & \textbf{56.46} & 27.18 & \textbf{49.80} \\
 & RAFT (Offline) & \textbf{38.83} & 65.21 & 56.49 & 52.47 & \textbf{29.44} & 48.49 \\
\bottomrule
\end{NiceTabular}}
\label{tab:raft_ablation}
\end{table*}

\subsection{Impact of Judge Model Capacity}

We investigated the sensitivity of our training framework to the capacity of the Judge model used to compute rewards. Table \ref{tab:judge_ablation} compares the performance when training with a 3B-parameter Judge versus a 7B-parameter Judge. The results indicate that a larger Judge generally provides a higher-fidelity signal, leading to better final RAG performance across the board.

\begin{table*}[p]
\small
\centering
\caption{Ablation on Judge Model Size. We report the downstream RAG F1 scores using \textbf{Qwen-7B} as the KG constructor. Training with a stronger Judge (7B) generally yields better downstream performance than a smaller Judge (3B).}
\resizebox{0.7\textwidth}{!}{
\begin{NiceTabular}{@{}l|c|cccccc@{}}
\toprule
\textbf{Retriever} & \textbf{Judge} & \textbf{NQ} & \textbf{PopQA} & \textbf{HotpotQA} & \textbf{2wiki} & \textbf{Musique} & \textbf{Avg} \\
\midrule
\multirow{2}{*}{Subgraph} & 3B & 27.10 & \textbf{61.76} & 39.71 & 37.31 & 14.87 & 36.15 \\
 & 7B & \textbf{28.54} & 60.94 & \textbf{43.59} & \textbf{37.43} & \textbf{15.65} & \textbf{37.23} \\
\midrule
\multirow{2}{*}{Triples} & 3B & \textbf{34.20} & \textbf{59.20} & \textbf{48.40} & 31.12 & 19.04 & 38.39 \\
 & 7B & 33.98 & 58.02 & 48.28 & \textbf{36.04} & \textbf{20.56} & \textbf{39.38} \\
\midrule
\multirow{2}{*}{ToG} & 3B & 28.95 & 62.86 & 43.68 & 47.43 & \textbf{20.05} & 40.59 \\
 & 7B & \textbf{29.36} & 62.85 & \textbf{44.68} & \textbf{50.20} & 19.31 & \textbf{41.28} \\
\bottomrule
\end{NiceTabular}}
\label{tab:judge_ablation}
\end{table*}

\subsection{Integration with Existing KG-RAG Pipelines}
To verify the transferability of our method, we integrated our fine-tuned Qwen-7B model into the official HippoRAG \citep{gutierrez_hipporag_2025} codebase, replacing their default zero-shot construction model. As shown in Table \ref{tab:hipporag_integration}, our model functions as a plug-and-play enhancement, improving both F1 and Recall metrics across datasets.
\begin{table*}[p]
\small
\centering
\caption{Performance of the HippoRAG KG Construction pipeline when replacing the default KG constructor with our fine-tuned Qwen-7B model. The consistent improvement confirms that our optimized KG construction transfers benefits to established third-party RAG systems.}
\resizebox{0.8\textwidth}{!}{
\begin{NiceTabular}{@{}l|cccccc@{}}
\toprule
\textbf{Method} & \textbf{NQ} & \textbf{PopQA} & \textbf{HotpotQA} & \textbf{2wiki} & \textbf{Musique} & \textbf{Avg} \\
\midrule
\rowcolor[gray]{0.9}\textit{HippoRAG F1 Score} \\
\midrule
Original (Base 7B) & 28.78 & 37.83 & 41.34 & 33.66 & 18.85 & 32.09 \\
\rowcolor{beaublue} + Our Fine-tuned 7B & 28.47 & \textbf{38.04} & \textbf{42.84} & 32.24 & \textbf{20.74} & \textbf{32.47} \\
\midrule
\rowcolor[gray]{0.9}\textit{HippoRAG Recall Score} \\
\midrule
Original (Base 7B) & 63.89 & 51.65 & 83.60 & 76.76 & 53.88 & 65.96 \\
\rowcolor{beaublue} + Our Fine-tuned 7B & 63.28 & \textbf{52.10} & \textbf{84.00} & \textbf{78.35} & \textbf{54.01} & \textbf{66.35} \\
\bottomrule
\end{NiceTabular}}
\label{tab:hipporag_integration}
\end{table*}

\subsection{Additional Results on Llama Models}
\label{sec:appendix_llama_result}

In this section, we present the performance improvements of AutoGraph-R1 when applied to the Llama-3.2 model family. These results (Tables \ref{tab:graph_retrievers_llama}, \ref{tab:graph_text_retrievers_llama}, and \ref{tab:recall_benchmarks_llama}) demonstrate that our method generalizes effectively across different LLM architectures.

\subsection{Structural Analysis of Constructed KGs}
To understand how our reward functions alter the topology of the generated graphs, we analyzed the structural metrics of KGs generated by the Qwen model family across four datasets: 2021Wiki, 2WikiMultihopQA, HotpotQA, and Musique.
Table \ref{tab:detailed_kg_statistics} presents the granular statistics for both 3B and 7B models across all evaluated datasets.
The analysis reveals two distinct, reward-driven strategies:

\begin{itemize} \item Optimizing for Completeness (Knowledge-Carrying Reward): The model trained for graph retrieval (Knowledge-Carrying) maintains a high number of triples but generates a dramatically larger vocabulary of unique relation types (e.g., +34$\%$ on average for Qwen-7B). This is direct evidence that the model learns to be more descriptive, discovering nuanced, long-tail relations instead of generic ones. This creates a richer graph structure necessary for multi-hop reasoning. \item Optimizing for Precision (Knowledge-Indexing Reward): Conversely, the model trained for the text retriever (Knowledge-Indexing) learns to be more concise. It produces fewer total triples per document (e.g., -12$\%$ on average for Qwen-7B) and a more focused set of relations. This demonstrates an optimization for efficiency, where the model actively filters out noise to create a cleaner, high-signal "index" for text retrieval. \end{itemize}

These results confirm that \textbf{AutoGraph-R1} is not a monolithic filter but a targeted optimization framework that reshapes the KG topology to maximize downstream utility.

\begin{table*}[p]
\small
\centering
\caption{Impact of our Knowledge-Carrying reward finetuned KG constructor on Graph Knowledge Retriever performance across the Llama model family.}
\resizebox{0.9\textwidth}{!}{
\begin{NiceTabular}{@{}l|llllll@{}}
\toprule
\multirow{2}{*}{Methods} & \multicolumn{2}{c}{Simple QA} & \multicolumn{3}{c}{Multihop QA} & \multicolumn{1}{c}{Overall} \\
\cmidrule{2-7}
& NQ* & PopQA* & HotpotQA & 2WikiMultihopQA & Musique & Avg. \\
\midrule
\rowcolor[gray]{0.9}{\textit{Llama-3.2-1B}} \\
\midrule
Subgraph (Base) & 23.17 & 26.14 & 27.11 & 19.36 & 9.13 & 20.98 \\
\rowcolor{beaublue}Subgraph (Ours) & \textbf{25.72}\textcolor{red}{$\uparrow_{2.55}$} & \textbf{47.32}\textcolor{red}{$\uparrow_{21.18}$} & \textbf{35.84}\textcolor{red}{$\uparrow_{8.73}$} & \textbf{25.05}\textcolor{red}{$\uparrow_{5.69}$} & \textbf{11.90}\textcolor{red}{$\uparrow_{2.77}$} & \textbf{29.17}\textcolor{red}{$\uparrow_{8.19}$} \\
Triples Retriever (Base) & 21.51 & 24.55 & 29.86 & 18.54 & 10.79 & 21.05 \\
\rowcolor{beaublue} Triples Retriever (Ours) & \textbf{24.96}\textcolor{red}{$\uparrow_{3.45}$} & \textbf{41.23}\textcolor{red}{$\uparrow_{16.68}$} & \textbf{36.25}\textcolor{red}{$\uparrow_{6.39}$} & \textbf{23.88}\textcolor{red}{$\uparrow_{5.34}$} & \textbf{13.89}\textcolor{red}{$\uparrow_{3.10}$} & \textbf{28.04}\textcolor{red}{$\uparrow_{6.99}$} \\
ToG (Base) & 20.77 & 18.17 & 25.63 & 15.29 & 7.79 & 17.53 \\
\rowcolor{beaublue}ToG (Ours) & 19.74 & \textbf{34.20}\textcolor{red}{$\uparrow_{16.03}$} & \textbf{30.70}\textcolor{red}{$\uparrow_{5.07}$} & \textbf{21.35}\textcolor{red}{$\uparrow_{6.06}$} & \textbf{9.60}\textcolor{red}{$\uparrow_{1.81}$} & \textbf{23.12}\textcolor{red}{$\uparrow_{5.59}$} \\
\hline
\midrule
\rowcolor[gray]{0.9}{\textit{Llama-3.2-3B}} \\
\midrule
Subgraph (Base) & 25.19 & 44.07 & 34.86 & 26.23 & 12.26 & 28.52 \\
\rowcolor{beaublue}Subgraph (Ours) & \textbf{27.34}\textcolor{red}{$\uparrow_{2.15}$} & \textbf{53.50}\textcolor{red}{$\uparrow_{9.43}$} & \textbf{40.89}\textcolor{red}{$\uparrow_{6.03}$} & \textbf{33.91}\textcolor{red}{$\uparrow_{7.68}$} & \textbf{15.50}\textcolor{red}{$\uparrow_{3.24}$} & \textbf{34.23}\textcolor{red}{$\uparrow_{5.71}$} \\
Triples Retriever (Base) & 26.18 & 43.80 & 39.11 & 27.03 & 15.21 & 30.27 \\
\rowcolor{beaublue}Triples Retriever (Ours) & \textbf{30.93}\textcolor{red}{$\uparrow_{4.75}$} & \textbf{51.50}\textcolor{red}{$\uparrow_{7.70}$} & \textbf{43.67}\textcolor{red}{$\uparrow_{4.56}$} & \textbf{31.49}\textcolor{red}{$\uparrow_{4.46}$} & \textbf{18.21}\textcolor{red}{$\uparrow_{3.00}$} & \textbf{35.16}\textcolor{red}{$\uparrow_{4.89}$} \\
ToG (Base) & 20.00 & 31.33 & 31.93 & 27.20 & 11.14 & 24.32 \\
\rowcolor{beaublue}ToG (Ours) & \textbf{23.78}\textcolor{red}{$\uparrow_{3.78}$} & \textbf{48.55}\textcolor{red}{$\uparrow_{17.22}$} & \textbf{40.60}\textcolor{red}{$\uparrow_{8.67}$} & \textbf{39.79}\textcolor{red}{$\uparrow_{12.59}$} & \textbf{17.31}\textcolor{red}{$\uparrow_{6.17}$} & \textbf{34.01}\textcolor{red}{$\uparrow_{9.69}$} \\
\hline
\bottomrule
\end{NiceTabular}}
\label{tab:graph_retrievers_llama}
\end{table*}

\begin{table*}[p]
\small
\centering
\caption{Impact of our Knowledge-Indexing reward finetuned KG constructor on Graph-based Text Retriever performance across the Llama model family.}
\resizebox{0.9\textwidth}{!}{
\begin{NiceTabular}{@{}l|llllll@{}}
\toprule
\multirow{2}{*}{Methods} & \multicolumn{2}{c}{Simple QA} & \multicolumn{3}{c}{Multihop QA} & \multicolumn{1}{c}{Overall} \\
\cmidrule{2-7}
& NQ* & PopQA* & HotpotQA & 2WikiMultihopQA & Musique & Avg. \\
\midrule
\rowcolor[gray]{0.9}{\textit{Llama-3.2-1B}} \\
\midrule
HippoRAG (Base) & 28.94 & 39.16 & 39.37 & 27.26 & 18.89 & 30.72 \\
\rowcolor{beaublue} HippoRAG (Ours) & \textbf{35.49}\textcolor{red}{$\uparrow_{6.55}$} & \textbf{52.63}\textcolor{red}{$\uparrow_{13.47}$} & \textbf{47.31}\textcolor{red}{$\uparrow_{7.94}$} & \textbf{38.13}\textcolor{red}{$\uparrow_{10.87}$} & \textbf{20.17}\textcolor{red}{$\uparrow_{1.28}$} & \textbf{38.75}\textcolor{red}{$\uparrow_{8.03}$} \\
HippoRAG2 (Base) & 29.79 & 40.64 & 37.90 & 29.82 & 20.21 & 31.67 \\
\rowcolor{beaublue} HippoRAG2 (Ours) & \textbf{35.61}\textcolor{red}{$\uparrow_{5.82}$} & \textbf{51.66}\textcolor{red}{$\uparrow_{11.02}$} & \textbf{50.97}\textcolor{red}{$\uparrow_{13.07}$} & \textbf{41.24}\textcolor{red}{$\uparrow_{11.42}$} & \textbf{21.58}\textcolor{red}{$\uparrow_{1.37}$} & \textbf{40.21}\textcolor{red}{$\uparrow_{8.54}$} \\
\hline
\midrule
\rowcolor[gray]{0.9}{\textit{Llama-3.2-3B}} \\
\midrule
HippoRAG (Base) & 38.35 & 58.81 & 50.68 & 44.26 & 23.63 & 43.15 \\
\rowcolor{beaublue}HippoRAG (Ours) & \textbf{38.77}\textcolor{red}{$\uparrow_{0.42}$} & \textbf{59.62}\textcolor{red}{$\uparrow_{0.81}$} & \textbf{54.82}\textcolor{red}{$\uparrow_{4.14}$} & \textbf{47.03}\textcolor{red}{$\uparrow_{2.77}$} & \textbf{25.26}\textcolor{red}{$\uparrow_{1.63}$} & \textbf{45.10}\textcolor{red}{$\uparrow_{1.95}$} \\
HippoRAG2 (Base) & 38.20 & 60.18 & 52.86 & 46.44 & 23.78 & 44.29 \\
\rowcolor{beaublue}HippoRAG2 (Ours) & \textbf{38.54}\textcolor{red}{$\uparrow_{0.34}$} & \textbf{61.05}\textcolor{red}{$\uparrow_{0.87}$} & \textbf{54.43}\textcolor{red}{$\uparrow_{1.57}$} & \textbf{48.79}\textcolor{red}{$\uparrow_{2.35}$} & 23.44 & \textbf{45.25}\textcolor{red}{$\uparrow_{0.96}$}\\
\hline
\bottomrule
\end{NiceTabular}}
\label{tab:graph_text_retrievers_llama}
\vspace{-0.1in}
\end{table*}

\begin{table*}[p]
\small
\centering
\caption{Evaluating Knowledge Indexing Quality via Passage Recall across the Llama model family.}
\label{tab:recall_benchmarks_llama}
\resizebox{0.9\textwidth}{!}{
\begin{NiceTabular}{@{}l|llllll@{}}
\toprule
\multirow{2}{*}{Methods} & \multicolumn{2}{c}{Simple QA} & \multicolumn{3}{c}{Multihop QA} & \multicolumn{1}{c}{Overall} \\
\cmidrule{2-7}
& NQ* & PopQA* & HotpotQA & 2WikiMultihopQA & Musique & Avg. \\
\midrule
\rowcolor[gray]{0.9} {\textit{Llama-3.2-1B}} \\
\midrule
HippoRAG (Base) & 57.50 & 51.30 & 37.86 & 36.78 & 24.35 & 41.56 \\
\rowcolor{beaublue}HippoRAG (Ours) & \textbf{66.30}\textcolor{red}{$\uparrow_{8.80}$} & \textbf{71.60}\textcolor{red}{$\uparrow_{20.30}$} & \textbf{56.56}\textcolor{red}{$\uparrow_{18.70}$} & \textbf{56.29}\textcolor{red}{$\uparrow_{19.51}$} & \textbf{33.99}\textcolor{red}{$\uparrow_{9.64}$} & \textbf{56.95}\textcolor{red}{$\uparrow_{15.39}$} \\
HippoRAG2 (Base) & 65.00 & 54.80 & 42.02 & 38.90 & 26.48 & 45.44 \\
\rowcolor{beaublue}HippoRAG2 (Ours) & \textbf{83.00}\textcolor{red}{$\uparrow_{18.00}$} & \textbf{73.70}\textcolor{red}{$\uparrow_{18.90}$} & \textbf{60.51}\textcolor{red}{$\uparrow_{18.49}$}  &\textbf{ 58.19}\textcolor{red}{$\uparrow_{19.29}$} & \textbf{36.44}\textcolor{red}{$\uparrow_{9.96}$} & \textbf{62.37}\textcolor{red}{$\uparrow_{16.93}$} \\
\hline
\midrule
\rowcolor[gray]{0.9} {\textit{Llama-3.2-3B}} \\
\midrule
HippoRAG (Base) & 87.70 & 85.40 & 60.80 & 63.97 & 39.02 & 67.38 \\
\rowcolor{beaublue}HippoRAG (Ours) & \textbf{90.90}\textcolor{red}{$\uparrow_{3.20}$} & \textbf{88.50}\textcolor{red}{$\uparrow_{3.10}$} & \textbf{66.23}\textcolor{red}{$\uparrow_{5.43}$} & \textbf{68.77}\textcolor{red}{$\uparrow_{4.80}$} & \textbf{41.63}\textcolor{red}{$\uparrow_{2.61}$} & \textbf{71.21}\textcolor{red}{$\uparrow_{3.83}$} \\
HippoRAG2 (Base) & 89.70 & 86.70 & 63.85 & 66.42 & 40.69 & 69.47 \\
\rowcolor{beaublue}HippoRAG2 (Ours) & \textbf{91.20}\textcolor{red}{$\uparrow_{1.50}$} & \textbf{89.00}\textcolor{red}{$\uparrow_{2.30}$} & \textbf{68.00}\textcolor{red}{$\uparrow_{4.15}$} & \textbf{71.57}\textcolor{red}{$\uparrow_{5.15}$} & \textbf{43.23}\textcolor{red}{$\uparrow_{2.54}$} & \textbf{72.60}\textcolor{red}{$\uparrow_{3.13}$} \\
\hline
\bottomrule
\end{NiceTabular}}
\vspace{-0.1in}
\end{table*}

\section{Impact of Using F1 Reward on Training Dynamics of AutoGraph-R1}
\label{sec:f1_reward_autograph-r1}
To validate our choice of using task-specific rewards ($R_C$ and $R_I$), we conducted an ablation study comparing them against the final answer's F1 score as a direct reward signal. While the downstream performance results are reported in Table~\ref{tab:ablation_unified}, here we analyze the training dynamics to further explain why the F1 reward is an inferior choice for guiding graph construction.

\begin{figure}[!h]
    \centering
    \begin{subfigure}[b]{0.48\textwidth}
        \centering
        \includegraphics[width=\linewidth]{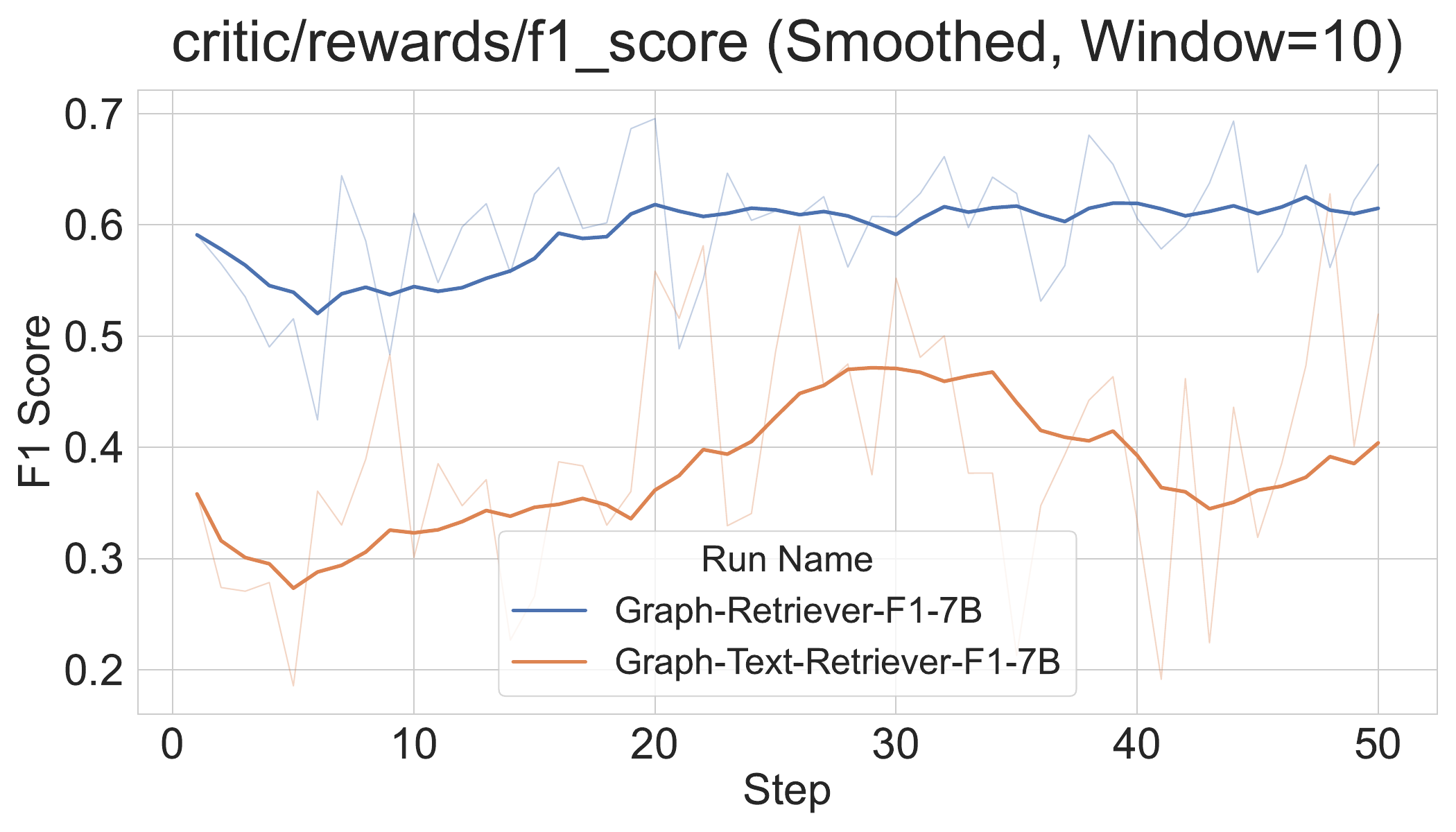}
        \caption{Volatile reward signal.}
        \label{fig:f1_reward_curve}
    \end{subfigure}
    \hfill
    \begin{subfigure}[b]{0.48\textwidth}
        \centering
        \includegraphics[width=\linewidth]{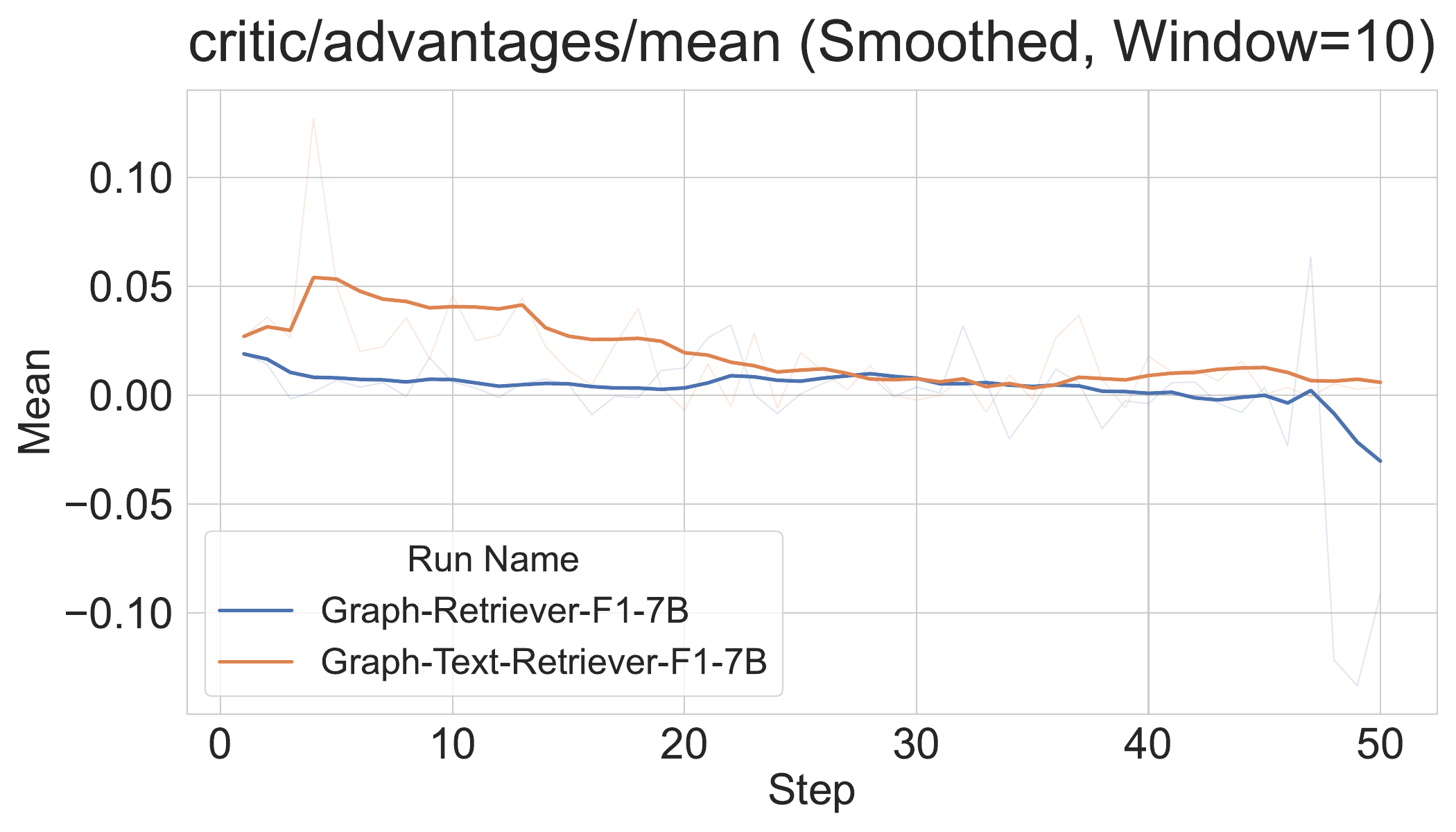}
        \caption{Stagnant advantage gain.}
        \label{fig:f1_adv_curve}
    \end{subfigure}
    
    \caption{\textbf{Unstable Training Dynamics Using a Naive F1 Reward.} Training curves for the ablation study where the final RAG F1 score is used as the reward. (a) The F1 reward signal is highly volatile and shows no clear upward trend, providing a noisy and ineffective learning signal. (b) Consequently, the advantage gain remains flat and centered around zero, confirming that the policy is failing to find a consistent direction for improvement. This leads to stalled optimization, as reflected in the poor downstream results.}
    \label{fig:dynamics_f1_ablation}
\end{figure}

\textbf{\textit{F1-based RL leads to unstable training}} Figure \ref{fig:f1_reward_curve} illustrates the instability inherent in using the final F1 score as a reward. The reward curve (Figure \ref{fig:f1_reward_curve}) exhibits high variance and lacks a clear, monotonic upward trend compared with using task specific reward, indicating a noisy learning signal.

\section{Training Dynamics of AutoGraph-R1}
\label{sec:dynamics_autograph-r1}
\begin{figure*}[p]
\vspace{-0.35in}
    \centering
    \begin{subfigure}[b]{0.45\textwidth}
        \centering
        \includegraphics[width=\linewidth]{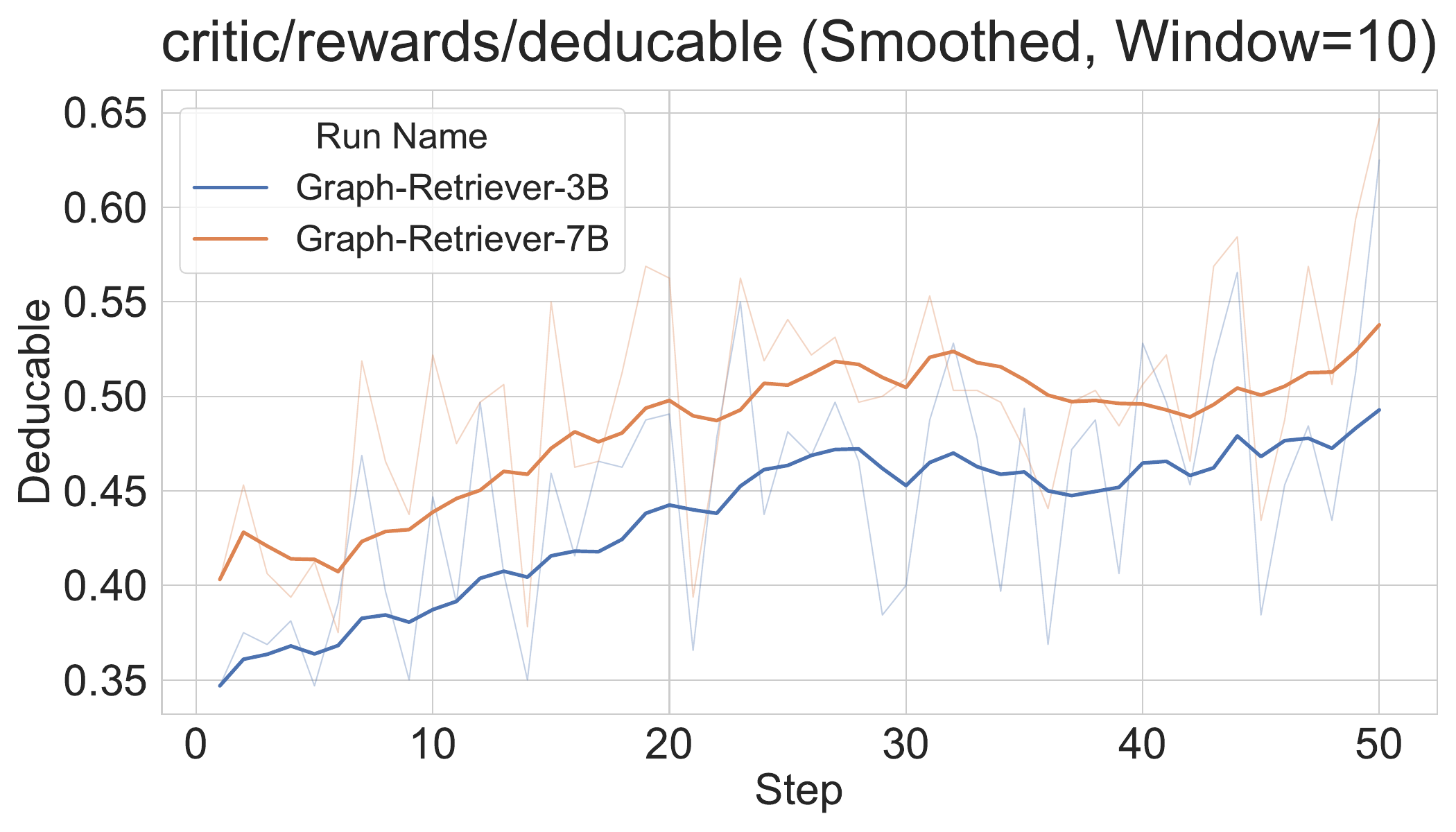}
        \caption{Reward convergence.}
        \label{fig:deducible_reward_curve}
    \end{subfigure}
    \hfill 
    \begin{subfigure}[b]{0.45\textwidth}
        \centering
        \includegraphics[width=\linewidth]{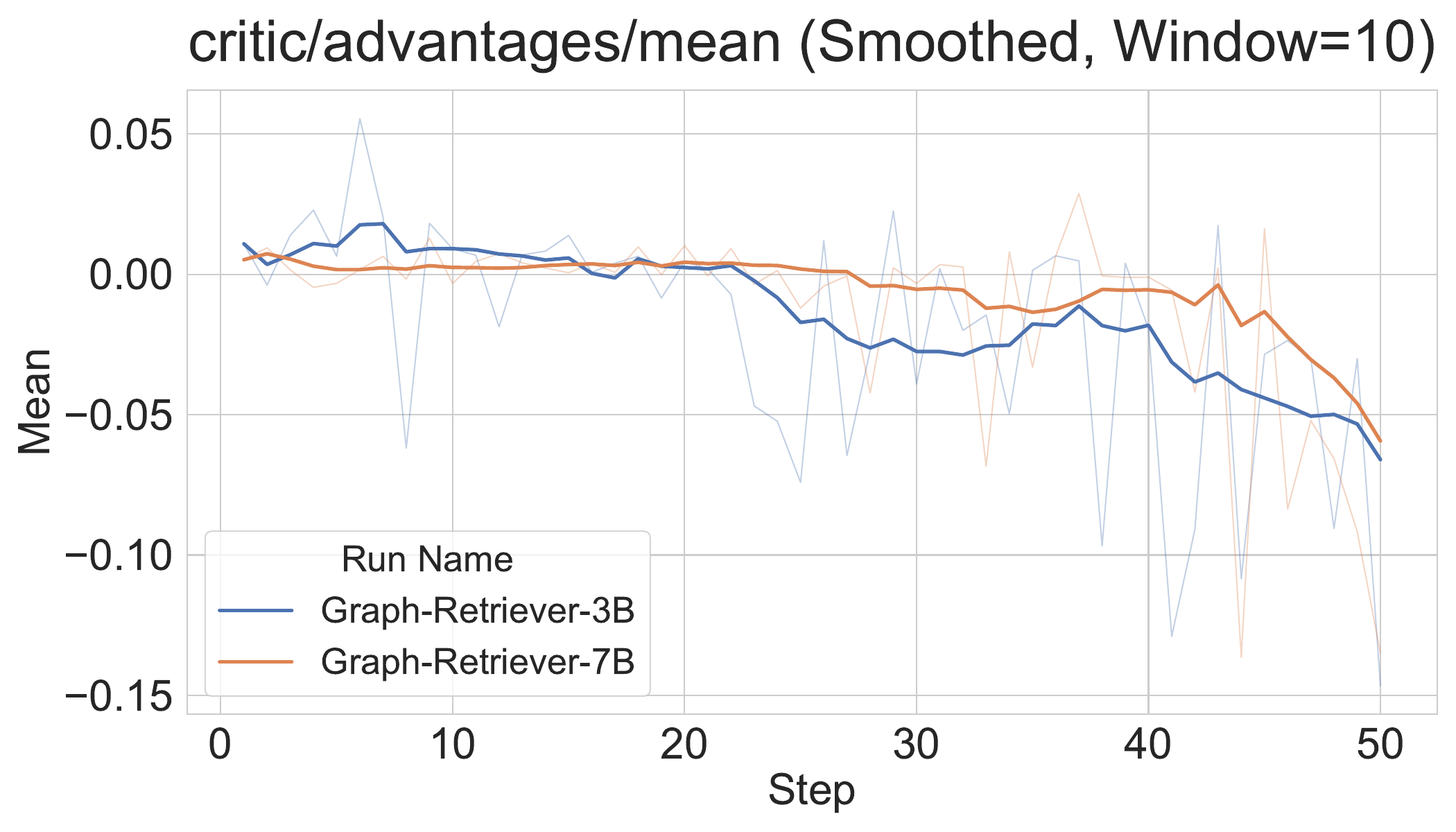}
        \caption{Positive advantage gain.}
        \label{fig:deducible_adv_curve}
    \end{subfigure}
    
    \caption{\textbf{Effective Training Dynamics with the Deducible Reward ($R_C$).} Training curves for the Graph Knowledge Retriever setting. (a) The reward, measuring answer deducibility, steadily increases and converges, demonstrating the policy is successfully learning its objective. (b) The advantage gain trends towards a small negative value, indicating that the value function's estimate of expected reward is rising quickly while the policy makes stable, incremental improvements. This dynamic, coupled with the rising absolute reward, points to effective and controlled optimization.}
    \label{fig:dynamics_deducible}
\end{figure*}

\begin{figure*}[p]
\vspace{-0.35in}
    \centering
    \begin{subfigure}[b]{0.45\textwidth}
        \centering
        \includegraphics[width=\linewidth]{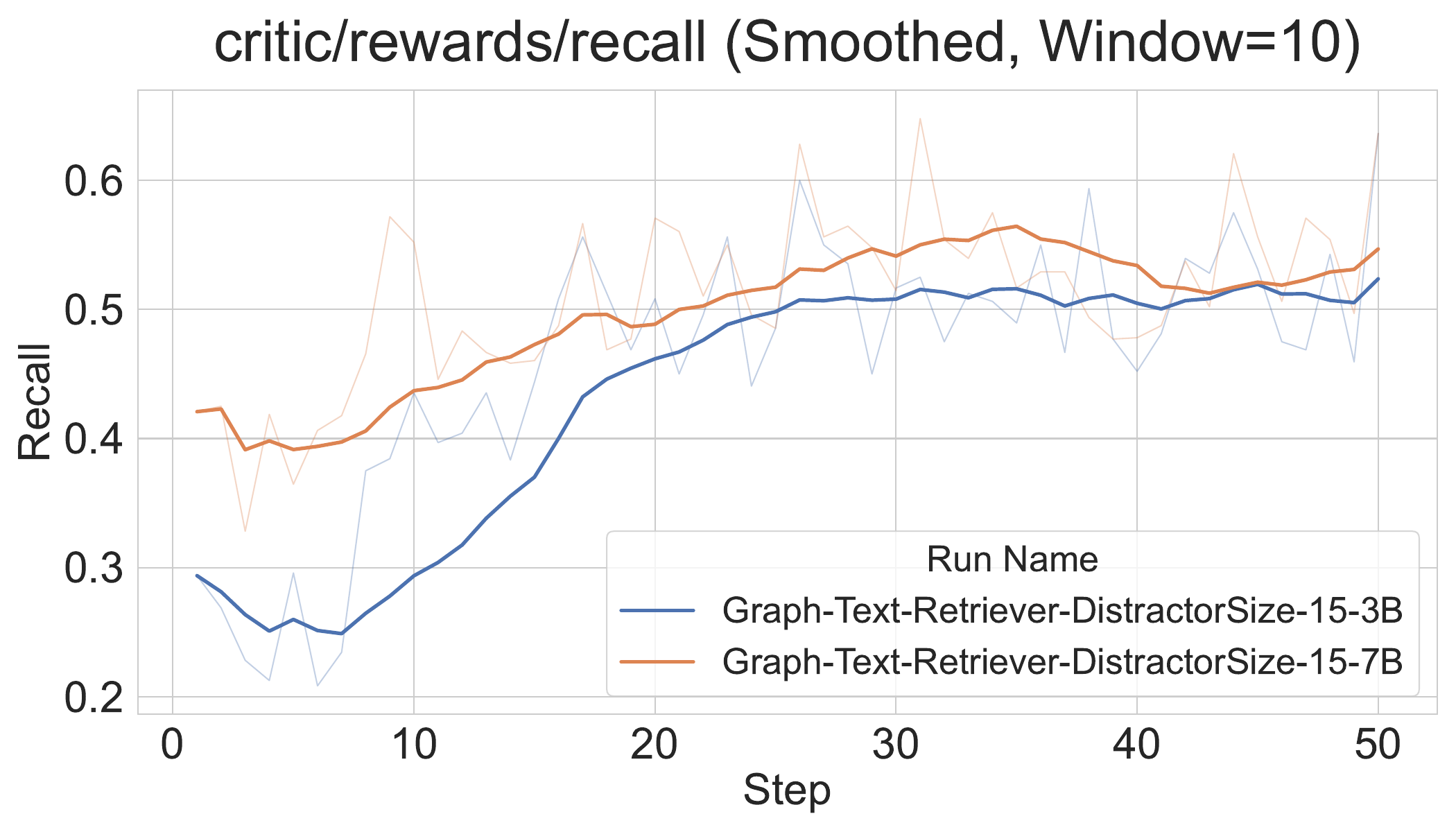}
        \caption{Reward convergence.}
        \label{fig:indexing_reward_curve}
    \end{subfigure}
    \hfill
    \begin{subfigure}[b]{0.45\textwidth}
        \centering
        \includegraphics[width=\linewidth]{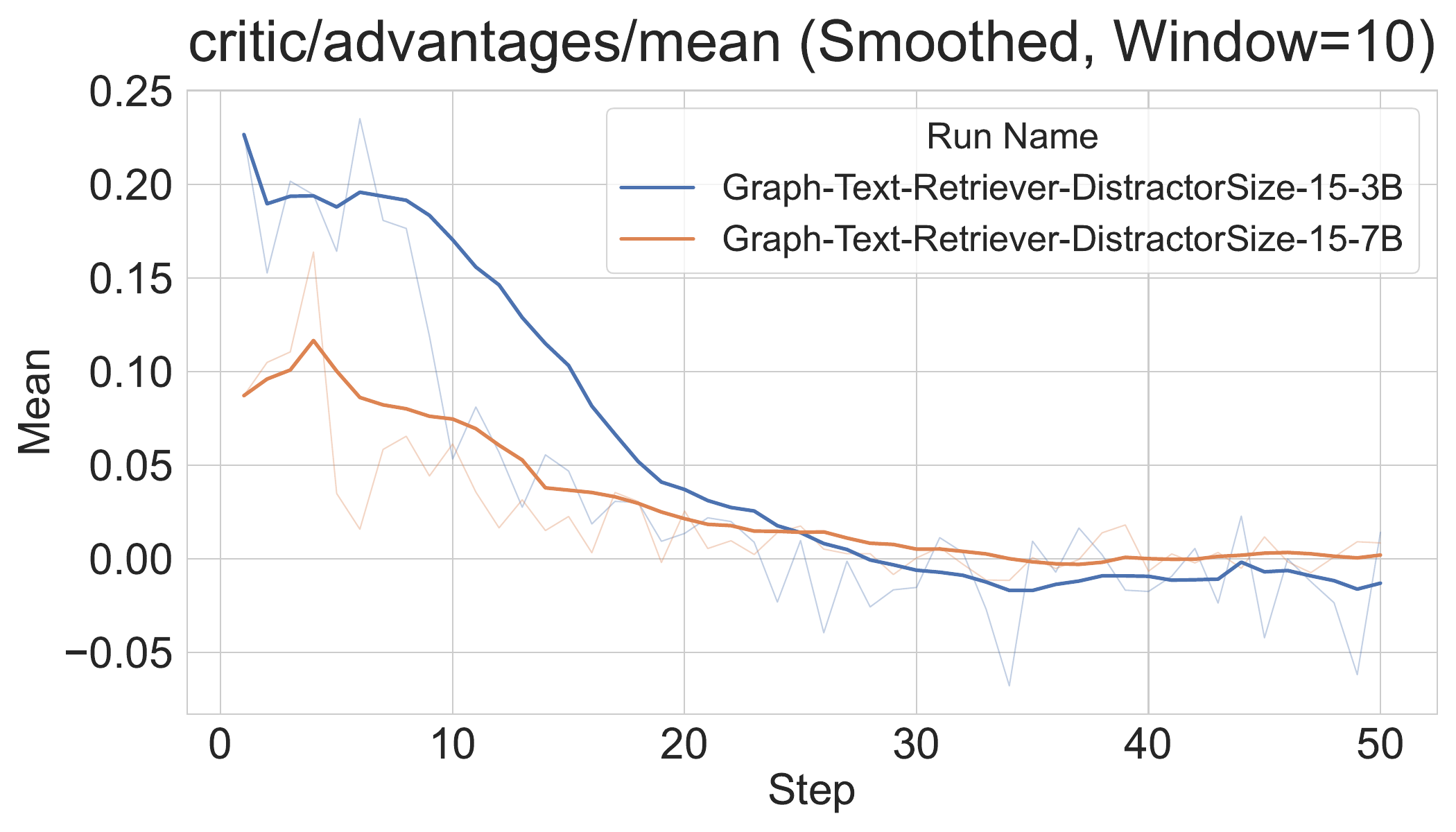}
        \caption{Positive advantage gain.}
        \label{fig:indexing_adv_curve}
    \end{subfigure}
    
    \caption{\textbf{Effective Training Dynamics with the Knowledge-Indexing Reward ($R_I$).} Training curves for the Graph-based Text Retriever setting. (a) The reward, measuring passage recall, shows a clear upward trend of improvement. (b) The advantage gain dynamic, paired with the rising reward curve, confirms that the policy is effectively learning from this stable, task-specific signal.}
    \label{fig:dynamics_indexing}
\end{figure*}

To verify the stability and effectiveness of our RL framework, we analyze the learning curves for both retrieval scenarios. Figure \ref{fig:dynamics_deducible} illustrates the training process for the Graph Knowledge Retriever using the Deducible Reward ($R_C$). We observe a steady increase in the reward signal, indicating that the model progressively learns to construct graphs with higher deductive validity. Similarly, Figure \ref{fig:dynamics_indexing} depicts the dynamics for the Graph-based Text Retriever using the Knowledge-Indexing Reward ($R_I$). The clear upward trend in recall-based reward, coupled with stable advantage estimates, confirms that AutoGraph-R1 successfully optimizes the graph structure for indexing utility without suffering from the volatility often associated with sparse RL rewards.
\section{Training Dynamics of Llama Models}
\label{sec:appendix_llama}
\begin{figure*}[p]
\centering
\begin{subfigure}[b]{0.45\textwidth}
\centering
\includegraphics[width=\linewidth]{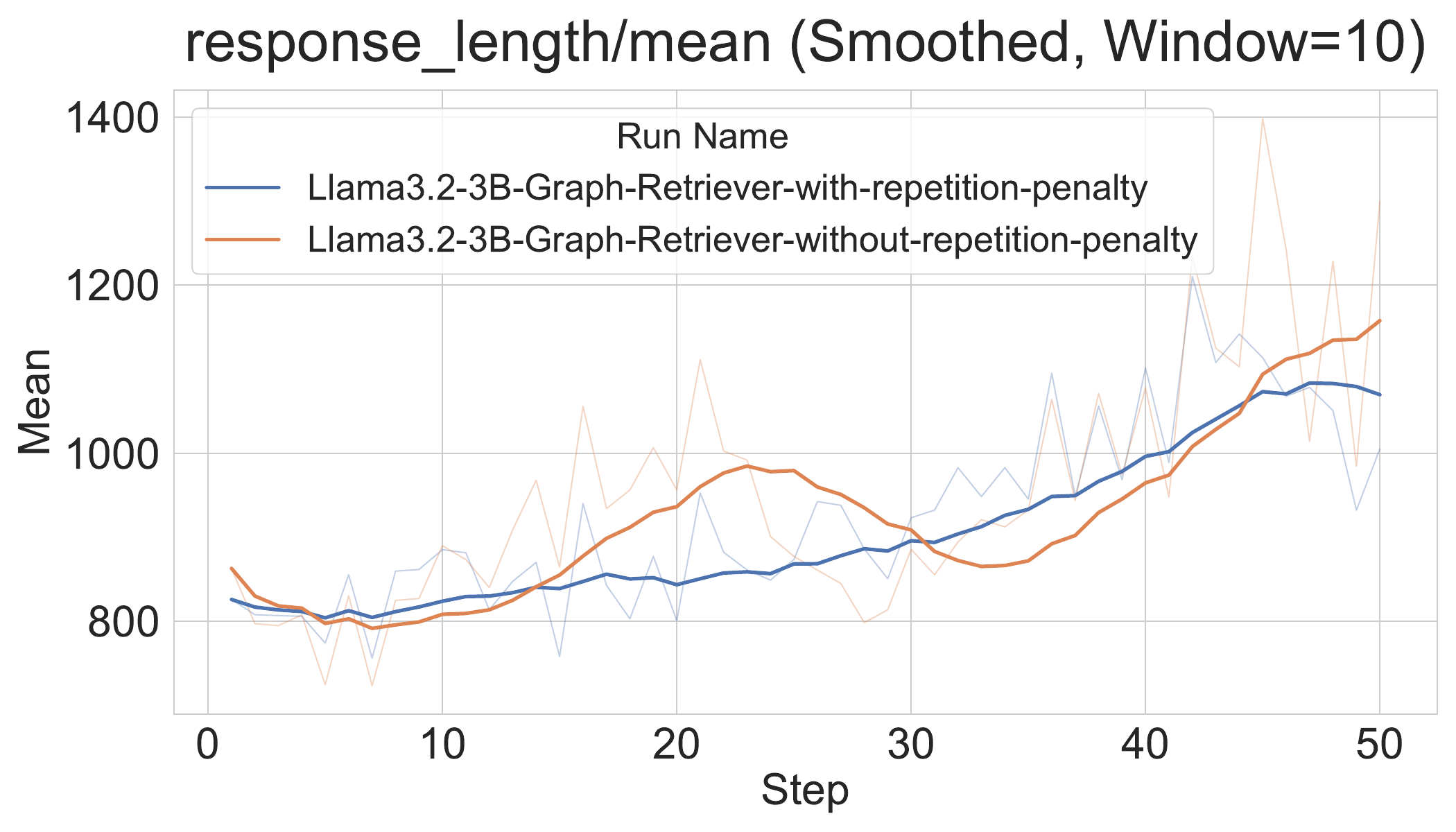}
\caption{Mean response length during training.}
\label{fig:llama_resp_mean}
\end{subfigure}
\hfill
\begin{subfigure}[b]{0.45\textwidth}
\centering
\includegraphics[width=\linewidth]{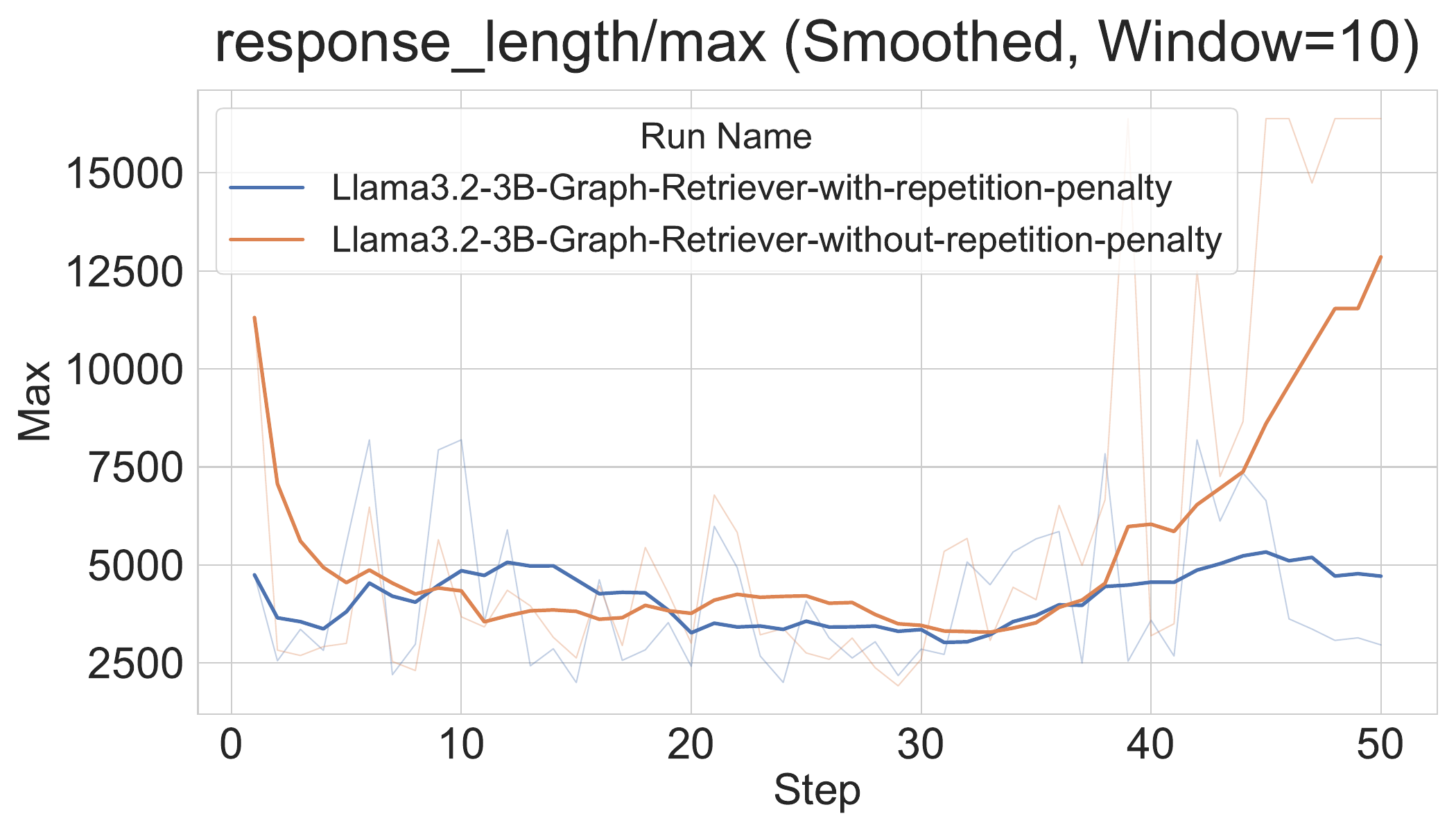}
\caption{Maximum response length during training.}
\label{fig:llama_resp_max}
\end{subfigure}
\caption{\textbf{Response Length Dynamics of Llama Models.} The Llama models showed unstable response lengths due to repetition issues. (a) Mean response length fluctuated significantly. (b) Maximum response length often reached the context window limit, indicating severe repetition loops.}
\label{fig:dynamics_llama_resp}
\vspace{-0.15em}
\end{figure*}

Our initial experiments with the Llama model series showed a tendency to generate repetitive triples, leading to unstable training dynamics. As seen in Figure \ref{fig:dynamics_llama_resp}, the model's generation length became erratic, with the mean response length fluctuating and the maximum often saturating the context window, indicating severe repetition loops.

To mitigate this issue, we augmented the reward for Llama experiments with a repetition penalty. The modified reward is defined as $\text{Reward} = R_{C/I} - \lambda_{\text{rep}} \cdot 
P_{\text{rep}}$, where $\lambda_{\text{rep}}$ is a scaling hyperparameter controlling 
the strength of the penalty, and the repetition penalty is calculated as 
$P_{\text{rep}} = (|T_{\text{gen}}| - |T_{\text{unique}}|) / |T_{\text{gen}}|$. Furthermore, we applied a hard constraint that sets the reward to 0 if $P_{\text{rep}} > 0.3$. This approach effectively discourages repetition, encouraging the model to produce diverse and high-quality facts.


\newpage
\section{Case Studies: The Functional Advantage of AutoGraph-R1}
\label{sec:appendix_case_study}
To qualitatively illustrate the benefits of our task-aware optimization, we present two case studies from the 2WikiMultiHopQA dataset. These examples demonstrate the \textbf{adaptive nature} of the utility objective: whereas Figure \ref{fig:teaser} showed the model learning to \textit{establish concrete pathways}, these case studies highlight its ability to \textit{bridge} disconnected reasoning chains. Together, they confirm that AutoGraph-R1 does not simply maximize or minimize graph size, but actively constructs the specific topology required for the downstream reasoning task.

\subsection{Case Study 1: Comparative Reasoning}
The first case study examines a question requiring a comparison between the death dates of two film directors. This task requires the KG to contain specific, comparable facts (i.e., dates) for multiple entities. As shown in Figure \ref{fig:case_study_comparison}, the zero-shot KG fails because it does not extract the specific death dates needed for comparison. In contrast, the KG constructed by AutoGraph-R1 contains the necessary date information, as the RL training has taught the constructor that dates are critical for such questions. This complete evidence enables the LLM to easily answer the question correctly.

\subsection{Case Study 2: Path-Based Reasoning}
The second case study involves a 2-hop question that requires finding a path from a film to its director, and then from the director to their child. This task depends on the structural connectivity of the graph.

As shown in Figure \ref{fig:case_study_path}, the zero-shot KG (top) fails critically. While it successfully extracts the first link in the path—` (Los Pagares de Mendieta, directed by, Leopoldo Torres Ríos)`—it fails to extract the second, crucial link about the director's child. The reasoning path is broken after the first hop, causing the QA system to fail. In contrast, the AutoGraph-R1 KG (bottom) explicitly contains the complete 2-hop reasoning path. It successfully extracts both `(Los Pagares de Mendieta, directed by, Leopoldo Torres Ríos)` and `(Leopoldo Torres Ríos, father of, Leopoldo Torre Nilsson)`. The RL process has rewarded the constructor for building these essential connective trails, recognizing that entity linkage across different relationships is crucial for multi-hop QA.

\begin{figure}[t]
    \centering
    \resizebox{0.9\columnwidth}{!}{
        \begin{minipage}{\linewidth} 
            \small
            \vspace{-0.2in}
            \begin{tcolorbox}[colback=black!5!white,colframe=black!75!black,title=Case Study 1: Zero-Shot KG (ToG Retriever) - \textbf{Failed Answer}]
                \textbf{Question:} Which film has the director who died first, The Goose Woman or You Can No Longer Remain Silent? \\
                \textbf{Retrieved Triples:} \\
                \texttt{"(You Can No Longer Remain Silent, directed by, Robert A. Stemmle)",} \\
                \texttt{"(Robert A. Stemmle, died in, Baden-Baden, Germany)",} \\
                \texttt{"(The Goose Woman, directed by, Clarence Brown)",} \\
                \texttt{"(Clarence Brown, was a, American film director)"} \\
                ... (and other irrelevant triples)
            \end{tcolorbox}
            \begin{tcolorbox}[colback=blue!5!white,colframe=blue!75!black,title=Case Study 1: AutoGraph-R1 KG (ToG Retriever) - \textbf{Correct Answer}]
                \textbf{Question:} Which film has the director who died first, The Goose Woman or You Can No Longer Remain Silent? \\
                \textbf{Retrieved Triples:} \\
                \texttt{"(You Can No Longer Remain Silent, directed by, Robert A. Stemmle)",} \\
                \texttt{\textbf{"(Robert A. Stemmle, died on, 24 February 1974)"}}, \\
                \texttt{"(The Goose Woman, directed by, Clarence Brown)",} \\
                \texttt{\textbf{"(Clarence Brown, died on, August 17, 1987)"}}, \\
                ...
            \end{tcolorbox}

            \caption{Qualitative comparison for a \textbf{comparative reasoning} question. The zero-shot KG lacks specific death dates, leading to failure. The AutoGraph-R1 KG, optimized for task utility, successfully extracts the critical dates needed for comparison.}
            \label{fig:case_study_comparison}
            
            \vspace{0.2in} 
            
            \begin{tcolorbox}[colback=black!5!white,colframe=black!75!black,title=Case Study 2: Zero-Shot KG (ToG Retriever) - \textbf{Failed Answer}]
                \textbf{Question:} Who is the child of the director of film Los Pagares De Mendieta? \\
                \textbf{Retrieved Triples:} \\
                \texttt{"(Los Pagares de Mendieta, directed by, Leopoldo Torres R\u00edos)",} \\ 
                \texttt{"(Leopoldo Torres R\u00edos, age at death, 60)",} \\
                ... (and other facts about the director, but not their child)
            \end{tcolorbox}
            \begin{tcolorbox}[colback=blue!5!white,colframe=blue!75!black,title=Case Study 2: AutoGraph-R1 KG (ToG Retriever) - \textbf{Correct Answer}]
                \textbf{Question:} Who is the child of the director of film Los Pagares De Mendieta? \\
                \textbf{Retrieved Triples:} \\
                \texttt{\textbf{"(Los Pagares de Mendieta, directed by, Leopoldo Torres R\u00edos)"}}, \\
                \texttt{\textbf{"(Leopoldo Torres R\u00edos, father of, Leopoldo Torre Nilsson)"}}, \\
                ...
            \end{tcolorbox}

            \caption{Qualitative comparison for a \textbf{2-hop path-based} question. The zero-shot KG extracts the first link (director of the film) but misses the second (child of the director), breaking the reasoning path. The AutoGraph-R1 KG successfully constructs the full path.}
            \label{fig:case_study_path}
            
        \end{minipage}
    }
\end{figure}

\clearpage

\section{AutoGraph-R1 Training Algorithm}
\label{sec:appendix_algorithm}
The end-to-end training process for the AutoGraph-R1 KG constructor is formalized in Algorithm \ref{alg:autograph}. The core idea is to iteratively construct a knowledge graph for a given query and its context documents, evaluate the graph's utility using a task-specific reward function, and then update the constructor's policy using the collected rewards.

\section{Prompts}
\label{sec:appendix_prompts}

This section details the specific prompts used in our experimental pipeline. The process begins with the graph construction prompt (Figure \ref{fig:prompt_kg_construction}), which guides the LLM to extract triples from raw text. During RL training, the \textbf{Knowledge-Carrying Reward} ($R_C$) is determined using the deducibility judge prompt shown in Figure \ref{fig:prompt_deducibility_judge}. For the final RAG answer generation step, we use distinct prompts tailored to the retrieved context: one for linearized graph triples (Figure \ref{fig:prompt_ans_gen_graph}) and another for raw text passages (Figure \ref{fig:prompt_ans_gen_text}). Finally, Figure \ref{fig:prompt_intrinsic_eval} shows the prompts used for our intrinsic graph quality analysis, where an LLM judge generates and answers multiple-choice questions to evaluate factual coverage.
\begin{figure}[H]
    \small
    \begin{AIbox}{{Multiple-Choice Question Generation and Answering}}
\textbf{MCQ Generation Prompt:}

You are an expert in generating multiple-choice questions (MCQs) from scientific texts.
Your task is to generate 5 multiple-choice questions based on the following passage.

Each question should:

\quad - Focus on factual claims, numerical data, definitions, or relational knowledge from the passage.

\quad - Have 4 options (one correct answer and three plausible distractors).

\quad - Clearly indicate the correct answer.

The output should be in JSON format, with each question as a dictionary containing:

\quad - "question": The MCQ question.

\quad - "options": A list of 4 options (e.g., ["A: ..", "B: ..", "C: ..", "D: .."]).

\quad - "answer": The correct answer (e.g., "A").

Passage:
\{passage\}

\hrulefill

\textbf{MCQ Answering Prompt:}

Given the contexts or evidences:
\{contexts\}

Here is a multiple-choice question:

Question: \{question\}

Options:
A. \{options\_0\}
B. \{options\_1\}
C. \{options\_2\}
D. \{options\_3\}

Please select the correct answer by choosing A, B, C, or D. Respond with only the letter of your choice.
    \end{AIbox}
    \caption{The prompt provided to the LLM judge (\texttt{DeepSeek-V3}) to evaluate triples extraction quality}
    \label{fig:prompt_intrinsic_eval}
\end{figure}
\begin{figure}[!ht]
    \small
    \begin{AIbox}{{Graph Construction}}
\textbf{Graph Generation System Prompt:}

You are an expert knowledge graph constructor.  
Your task is to extract factual information from the provided text and represent it strictly as a JSON array of knowledge graph triples.  

Output Format

\quad - The output must be a **JSON array**.

\quad - Each element in the array must be a **JSON object** with exactly three non-empty keys:

\quad \quad   - "subject": the main entity, concept, event, or attribute.  

\quad \quad   - "relation": a concise, descriptive phrase or verb that describes the relationship (e.g., "founded by", "started on", "is a", "has circulation of").  

\quad \quad   - "object": the entity, concept, value, event, or attribute that the subject has a relationship with.  

Constraints

\quad - **Do not include any text other than the JSON output.**

\quad - Do not add explanations, comments, or formatting outside of the JSON array.

\quad - Extract **all possible and relevant triples**.

\quad - All keys must exist and all values must be non-empty strings.

\quad - The "subject" and "object" can be specific entities (e.g., "Radio City", "Football in Albania", "Echosmith") or specific values (e.g., "3 July 2001", "1,310,696").

\quad - If no triples can be extracted, return exactly: `[]`.

Extracts for:
\{passage\}

    \end{AIbox}
    \caption{The prompt used for both zero-shot KG construction and fine-tuning KG constructor model during RL.}
    \label{fig:prompt_kg_construction}
\end{figure}

\begin{figure}[!ht]
    \small
    \begin{AIbox}{{Deducible Judge}}
\textbf{Deducible Judge Prompt:}

As an advanced reading comprehension assistant, your task is to evaluate whether the provided knowledge graph (KG) context contains sufficient information to deduce the given true answer to the question. 
Analyze the KG context carefully and determine if it fully supports the true answer without requiring external knowledge. Respond with only 'Yes' or 'No', indicating whether the true answer can be deduced from the KG context.

Knowledge graph (KG) context:\{triples string\} 

Question:\{query\}

True Answer:\{answer\}

Can the true answer be deduced from the KG context? Answer 'Yes' or 'No' only.

\end{AIbox}
\caption{The prompts for freeze LLM to determine the \textbf{Knowledge-Carrying Reward} ($R_C$). The 'Yes' or 'No' response serves as the binary reward signal.}
\label{fig:prompt_deducibility_judge}
\end{figure}

\begin{figure}[t!]
    \small
    \begin{AIbox}{{Graph Retriever Answer Generation}}
\textbf{Answer Generation Prompt For Graph Retriever:}

As an advanced reading comprehension assistant, your task is to analyze extracted information and corresponding questions meticulously. If the knowledge graph information is not enough, you can use your own knowledge to answer the question. 
Your response start after "Thought: ", where you will methodically break down the reasoning process, illustrating how you arrive at conclusions. 
Conclude with "Answer: " to present a concise, definitive response as a noun phrase, no elaborations.

\{triples string\}

\{question\}

Thought:
\end{AIbox}
\caption{The prompt used by the final answer generator when the retrieved evidence consists of linearized knowledge graph triples.}
\label{fig:prompt_ans_gen_graph}
\end{figure}

\begin{figure}[t!]
    \small
    \begin{AIbox}{{Graph Text Retriever Answer Generation}}
\textbf{Answer Generation Prompt:}

As an advanced reading comprehension assistant, your task is to analyze text passages and corresponding questions meticulously. If the information is not enough, you can use your own knowledge to answer the question.
Your response start after "Thought: ", where you will methodically break down the reasoning process, illustrating how you arrive at conclusions. 
Conclude with "Answer: " to present a concise, definitive response as a noun phrase, no elaborations.

\{Retrieved Texts\}

\{question\}

Thought:

\end{AIbox}
\caption{The prompt used by the final answer generator when the retrieved evidence consists of raw text passages.}
\label{fig:prompt_ans_gen_text}
\end{figure}
\vspace{-1in}
\begin{algorithm*}[b!]
\caption{AutoGraph-R1 Training Loop}
\label{alg:autograph}
\begin{algorithmic}[1]
\State \textbf{Input:} Training dataset $\mathcal{S} = \{(q_i, y_i, \mathcal{D}_{q_i})\}_{i=1}^N$, where $\mathcal{D}_{q_i}$ are the context documents for query $q_i$.
\State \textbf{Input:} KG constructor policy $\pi_{\theta}^{KG}$ (an LLM).
\State \textbf{Input:} Frozen retriever $\mathcal{R}_{\text{frozen}}$ (either a graph knowledge retriever or a graph-based text retriever).
\State \textbf{Input:} Chosen reward function $R_{\text{task}}$ (either $R_C$ or $R_I$).
\State \textbf{Initialize:} Policy parameters $\theta$.

\For{each training step}
    \State Sample a minibatch of data $\{(q, y, \mathcal{D}_q)\}$ from $\mathcal{S}$.
    \State Initialize an empty list of trajectories `trajectories`.

    \For{each sample $(q, y, \mathcal{D}_q)$ in the minibatch}
        \State \Comment{\textbf{Step 1: Construct the Knowledge Graph}}
        \State Generate the graph by sampling from the policy: $\mathcal{G} \sim \pi_{\theta}^{KG}(\cdot \mid \mathcal{D}_q)$.
        
        \State \Comment{\textbf{Step 2: Determine Task-Specific Reward}}
        \If{$R_{\text{task}}$ is Knowledge-Carrying Reward ($R_C$)}
            \State Use the frozen retriever $\mathcal{R}_{\text{graph}}$ to get evidence $\mathcal{P}(q)$ from $\mathcal{G}$.
            \State Calculate reward $r = R_C(q, y, \mathcal{P}(q))$ using Eq. (\ref{deducible_reward}).
        \ElsIf{$R_{\text{task}}$ is Knowledge-Indexing Reward ($R_I$)}
            \State Use the frozen retriever $\mathcal{R}_{\text{text}}$ to get passages $\mathcal{T}(q)$ from $\mathcal{G}$.
            \State Calculate reward $r = R_I(q, y, \mathcal{T}(q))$ using Eq. (\ref{recall_reward}).
        \EndIf
        
        \State Store the generation trajectory (actions taken to build $\mathcal{G}$) and the final reward $r$ in `trajectories`.
    \EndFor

    \State \Comment{\textbf{Step 3: Update Policy Parameters}}
    \State Compute the policy gradient $\nabla_{\theta} J(\theta)$ using the stored `trajectories` and a policy optimization algorithm (e.g., GRPO).
    \State Update the policy parameters: $\theta \leftarrow \theta - \eta \cdot \nabla_{\theta} J(\theta)$.
\EndFor
\State \textbf{Return:} Optimized KG constructor parameters $\theta$.
\end{algorithmic}
\end{algorithm*}

\begin{table*}[ht]
\scriptsize 
\centering
\caption{Detailed statistics of Knowledge Graphs constructed by different models across four benchmarks. The table tracks the total count of extracted entities, relations, triples, the density of extraction (Triples/Doc), and the diversity of the relation vocabulary (Unique Relation Types).}
\resizebox{0.9\linewidth}{!}{
\begin{NiceTabular}{@{}l|l|ccccc@{}}
    \toprule
    \textbf{Dataset} & \textbf{Model Variant} & \textbf{Total Ent.} & \textbf{Total Rel.} & \textbf{Total Trip.} & \textbf{Trip./Doc} & \textbf{Unique Rel.} \\
    \midrule
    \multirow{6}{*}{\textbf{2021Wiki}} 
    & Qwen-3B (Base) & 57,369 & 56,908 & 56,908 & 8.13 & 20,879 \\
    & \cellcolor{beaublue} Qwen-3B (Knwl-Carrying) & \cellcolor{beaublue} 67,660 & \cellcolor{beaublue} 65,413 & \cellcolor{beaublue} 65,413 & \cellcolor{beaublue} 9.34 & \cellcolor{beaublue} 28,933 \\
    & \cellcolor{beaublue} Qwen-3B (Knwl-Indexing) & \cellcolor{beaublue} 68,631 & \cellcolor{beaublue} 68,670 & \cellcolor{beaublue} 68,670 & \cellcolor{beaublue} 9.81 & \cellcolor{beaublue} 19,246 \\
    \cmidrule{2-7}
    & Qwen-7B (Base) & 75,521 & 70,264 & 70,264 & 10.04 & 23,887 \\
    & \cellcolor{beaublue} Qwen-7B (Knwl-Carrying) & \cellcolor{beaublue} 65,324 & \cellcolor{beaublue} 66,370 & \cellcolor{beaublue} 66,370 & \cellcolor{beaublue} 9.48 & \cellcolor{beaublue} 32,173 \\
    & \cellcolor{beaublue} Qwen-7B (Knwl-Indexing) & \cellcolor{beaublue} 72,529 & \cellcolor{beaublue} 61,324 & \cellcolor{beaublue} 61,324 & \cellcolor{beaublue} 8.76 & \cellcolor{beaublue} 22,201 \\
    \midrule
    \multirow{6}{*}{\textbf{2WikiMultihop}} 
    & Qwen-3B (Base) & 37,333 & 38,140 & 38,140 & 6.23 & 9,881 \\
    & \cellcolor{beaublue} Qwen-3B (Knwl-Carrying) & \cellcolor{beaublue} 44,520 & \cellcolor{beaublue} 43,623 & \cellcolor{beaublue} 43,623 & \cellcolor{beaublue} 7.13 & \cellcolor{beaublue} 13,177 \\
    & \cellcolor{beaublue} Qwen-3B (Knwl-Indexing) & \cellcolor{beaublue} 42,681 & \cellcolor{beaublue} 43,632 & \cellcolor{beaublue} 43,632 & \cellcolor{beaublue} 7.13 & \cellcolor{beaublue} 8,940 \\
    \cmidrule{2-7}
    & Qwen-7B (Base) & 45,740 & 46,074 & 46,074 & 7.53 & 11,362 \\
    & \cellcolor{beaublue} Qwen-7B (Knwl-Carrying) & \cellcolor{beaublue} 42,995 & \cellcolor{beaublue} 44,997 & \cellcolor{beaublue} 44,997 & \cellcolor{beaublue} 7.35 & \cellcolor{beaublue} 16,920 \\
    & \cellcolor{beaublue} Qwen-7B (Knwl-Indexing) & \cellcolor{beaublue} 44,675 & \cellcolor{beaublue} 40,192 & \cellcolor{beaublue} 40,192 & \cellcolor{beaublue} 6.57 & \cellcolor{beaublue} 10,609 \\
    \midrule
    \multirow{6}{*}{\textbf{HotpotQA}} 
    & Qwen-3B (Base) & 70,592 & 73,674 & 73,674 & 7.99 & 23,994 \\
    & \cellcolor{beaublue} Qwen-3B (Knwl-Carrying) & \cellcolor{beaublue} 77,593 & \cellcolor{beaublue} 76,943 & \cellcolor{beaublue} 76,943 & \cellcolor{beaublue} 8.34 & \cellcolor{beaublue} 30,816 \\
    & \cellcolor{beaublue} Qwen-3B (Knwl-Indexing) & \cellcolor{beaublue} 80,355 & \cellcolor{beaublue} 83,920 & \cellcolor{beaublue} 83,920 & \cellcolor{beaublue} 9.10 & \cellcolor{beaublue} 21,411 \\
    \cmidrule{2-7}
    & Qwen-7B (Base) & 82,253 & 81,237 & 81,237 & 8.81 & 24,996 \\
    & \cellcolor{beaublue} Qwen-7B (Knwl-Carrying) & \cellcolor{beaublue} 74,647 & \cellcolor{beaublue} 76,700 & \cellcolor{beaublue} 76,700 & \cellcolor{beaublue} 8.32 & \cellcolor{beaublue} 33,335 \\
    & \cellcolor{beaublue} Qwen-7B (Knwl-Indexing) & \cellcolor{beaublue} 79,861 & \cellcolor{beaublue} 71,598 & \cellcolor{beaublue} 71,598 & \cellcolor{beaublue} 7.76 & \cellcolor{beaublue} 23,677 \\
    \midrule
    \multirow{6}{*}{\textbf{Musique}} 
    & Qwen-3B (Base) & 77,462 & 78,988 & 78,988 & 6.78 & 29,255 \\
    & \cellcolor{beaublue} Qwen-3B (Knwl-Carrying) & \cellcolor{beaublue} 89,247 & \cellcolor{beaublue} 88,331 & \cellcolor{beaublue} 88,331 & \cellcolor{beaublue} 7.58 & \cellcolor{beaublue} 39,115 \\
    & \cellcolor{beaublue} Qwen-3B (Knwl-Indexing) & \cellcolor{beaublue} 91,223 & \cellcolor{beaublue} 93,930 & \cellcolor{beaublue} 93,930 & \cellcolor{beaublue} 8.06 & \cellcolor{beaublue} 27,132 \\
    \cmidrule{2-7}
    & Qwen-7B (Base) & 97,294 & 91,395 & 91,395 & 7.84 & 32,310 \\
    & \cellcolor{beaublue} Qwen-7B (Knwl-Carrying) & \cellcolor{beaublue} 89,547 & \cellcolor{beaublue} 92,605 & \cellcolor{beaublue} 92,605 & \cellcolor{beaublue} 7.94 & \cellcolor{beaublue} 40,078 \\
    & \cellcolor{beaublue} Qwen-7B (Knwl-Indexing) & \cellcolor{beaublue} 96,242 & \cellcolor{beaublue} 82,145 & \cellcolor{beaublue} 82,145 & \cellcolor{beaublue} 7.05 & \cellcolor{beaublue} 30,521 \\
    \bottomrule
\end{NiceTabular}}
\label{tab:detailed_kg_statistics}
\resizebox{0.5\linewidth}{!}{
\begin{NiceTabular}{@{}l|ccc@{}}
\toprule
\multicolumn{4}{c}{\textbf{Average Structural Metrics (Qwen-7B Family)}} \\
\midrule
\textbf{Metric} & \textbf{Base} & \textbf{Carrying} & \textbf{Indexing} \\
\midrule
Avg. Triples per Doc & 8.56 & 8.27 & \textbf{7.54} \\
Total Unique Rel. Types & 23,139 & \textbf{30,627} & 21,752 \\
\bottomrule
\end{NiceTabular}}

\end{table*}

\end{document}